\documentclass[ACS,STIX1COL]{WileyNJD-v2}

\usepackage[absolute]{overpic}
\usepackage{graphicx}

\articletype{Article Type}%

\received{xx xxx 2020}
\revised{xxx}
\accepted{xxx}

\begin{document}


\title{Machine Learning for Material Characterization with an Application for Predicting Mechanical Properties}

\author[1]{Anke Stoll}

\author[2]{Peter Benner}

\authormark{Stoll \textsc{et al}}

\address[1]{\orgdiv{Data-Mining and Machine Learning}, \orgname{Fraunhofer Institute for Machine Tools and Forming Technology IWU}, \orgaddress{\country{Chemnitz, Germany}}}

\address[2]{\orgdiv{Computational Methods in Systems and Control Theory}, \orgname{The Max Planck Institute for Dynamics of Complex Technical Systems}, \orgaddress{\country{Magdeburg, Germany}}}

\corres{*Anke Stoll \email{anke.stoll@iwu.fraunhofer.de}}

\presentaddress{Reichenhainer Stra\ss e 88, 09126 Chemnitz, Germany}

\abstract{Currently, the growth of material data from experiments and simulations is expanding beyond processable amounts. This makes the development of new data-driven methods for the discovery of patterns among multiple lengthscales and time-scales and structure-property relationships essential. These data-driven approaches
show enormous promise within materials science. The following review covers machine learning applications for metallic material characterization. Many parameters associated with the processing and the structure of materials affect the properties and the performance of manufactured components. Thus, this study is  an  attempt  to  investigate the  usefulness  of  machine learning  methods for material property prediction. Material characteristics such as strength, toughness, hardness, brittleness or ductility are relevant to categorize a material or component according to their quality. In industry, material tests like tensile tests, compression tests or creep tests are often time consuming and expensive to perform. Therefore, the application of machine learning approaches is considered helpful for an easier generation of material property information. This study also gives an application of machine learning methods on small punch test data  for the determination of the property ultimate tensile strength  for various materials. A strong correlation between small punch test data and  tensile test data was found which ultimately allows to replace more costly tests by simple and fast tests in combination with machine learning.}

\keywords{Machine Learning, Material Characterization, Small Punch Test, Ultimate Yield Strength, Tensile Properties}


\maketitle


\section{Introduction}\label{sec1}

The field of materials science relies on experiments and simulation-based models as tools for material characterization \cite{agrawal2016perspective}. Material properties, such as their structure and behaviour, are critical to the potential application of the material of interest. 
More recently, the data generated by such experiments and simulations have created  various opportunities for the application of data-driven methods.
In addition to, e.g., the experimental trial and error approach or a physical metallurgy approach, machine learning (ML) methods for property prediction and material design have attracted a lot of attention in recent years, see e.g. \cite{mueller2016machine, wagner2016theory, dimiduk2018perspectives}. 

While  experimental investigations (the so called first paradigm of materials science) have  been  carried  out  since  the  stone  and copper  age, scientists  of the 16th century started to describe physical relations by equations (second paradigm). Thus, analytical equations  became  a  central  instrument  of  theoretical  physics which were able to complement
the empirical and experimental sciences. The 1950s marked the  beginning  of  computational materials science and  simulations, the  third paradigm. Within this framework, computer experiments and simulations became possible,  with  the  corresponding  results being analyzed  and  interpreted  like measured ones. It had to be recognized  that  many properties of materials cannot be described by a closed mathematical form as they are  determined  by several  multi-level,  intricate theoretical  concepts. With the help of large amounts of data, hidden correlations, reflected in terms of structure and patterns in the data can be discovered that are not normally  visible  in  small  data  sets. Thus, the fourth paradigm,  data-driven science, of materials research was born \cite{draxl2018nomad, agrawal2016perspective}.

However, it is not only an advantage to have a large data volume but it can also be a challenge to cope with tremendous amounts of data. Today, data are indeed more and more easily acquired and stored, due to huge progresses in sensors and ways to collect data on one side, and in storage devices on the other side. Nowadays, there is  no  hesitation  in  many  domains  in  acquiring  very  large  amounts  of  data  without  knowing in advance if they will be analyzed and how. The spectacular increase in the amount of data is not only found in the number of samples  collected  for  example  over  time,  but  also  in  the  number  of  attributes,  or  characteristics,  that  are  simultaneously  measured  on  a  process.  Data  are  gathered  into  vectors  whose  dimension  correspond  to  the  number  of  simultaneous  measurements  on  the  process.  Growing  dimensions result in high dimensional data, as each sample can be represented as a point or vector in a high-dimensional space. Working  with  high-dimensional  data  means  working  with  data  that  are  embedded  in  high-dimensional  spaces\cite{verleysen2005curse} .The curse of dimensionality is the expression of all phenomena that appear  with  high-dimensional  data,  and  that  have  most  often  unfortunate  consequences on the behavior and performances of learning algorithms.  

Contrary to the curse of dimensionality, databases in materials science are often limited in size due to expensive and time consuming data acquisition via experiments or simulations\cite{shen2019physical}. Then the insufficient data size for the training of a ML model compromises the learning success and suitable new approaches for small datasets have to be found.

This work contains a literature survey which covers an overview of ML for materials science and specifically for metallic material characterization. As the measurement of such parameters is often expensive and time consuming obtained via experiments, alternative basic tests, such as the small punch test (SPT) can be an option if it can be shown that the same material property information can be extracted.

There is a wide range of ML approaches based on SPT data which will be presented. 
Furthermore, in Section 3 an example is described which uses ML for the prediction of  tensile properties of a insert-material-type based on SPT data. The objective of this study is to investigate whether it is possible to find a ML model which predicts/determines the tensile properties of a material from SPT data \cite{arunkumar2016estimation}. Section 4 concludes this paper by giving an outlook on further research perspectives.

\section{State of the Art}\label{sec2}
\subsection{Overview - machine learning for materials science} 

With ML, given enough data and a data-driven algorithm for rule discovery, a computer is able to determine physical laws which lead to the given data without human input\cite{butler2018machine, jordan2015machine}. Traditional computational approaches use the computer for the  employment of a hard-coded algorithm provided by a human expert. By contrast, ML approaches learn the rules that underlie a dataset by assessing a portion of that data and building a model to make predictions \cite{butler2018machine}. However, the human still needs to choose suitable ML models which supposedly represent the data well and do manual (sub-)tasks in pre-processing and feature generation.

The existence of large amounts of data makes the use of ML models possible and enables data-driven knowledge to be obtained and patterns to be discovered. On the other hand, big data and their high dimensionality  lead to difficult computational and statistical challenges, such as scalability and memory shortage, noise accumulation, interference correlation, incidental endogeneity and measurement errors \cite{fan2014challenges}. 

Materials science is an interesting field of application for big data methods and ML approaches which is beginning to show enormous promise. Four primary elements are critical in materials  science  and  engineering:  processing, structure,  properties,  and  performance \cite{national1974materials,olson1997computational}. There is no general agreement, however, on how these elements are interconnected. ML methods can be applied to the so called process-structure-property-performance chain for 
learning more about the intrinsic interrelations of these components. One main goal is the enabling, acceleration and simplification of the discovery and development
of novel materials  based on the convergence of high-performance computing, automation, and ML \cite{correa2018accelerating}. 
Another aim of using such approaches in the field of materials science is to achieve high-throughput identification and quantification of essential material properties \cite{bock2019review}.

Besides experimentally obtained datasets,
numerous studies draw required information from simulation-based data mining.
Altogether, it is shown that experiment- and simulation-based data mining in combination
with machine leaning tools provide exceptional opportunities to enable highly reliant
identification of fundamental interrelations within materials for characterization and optimization in a scale-bridging manner \cite{bock2019review}.

For more detailed information on recent ML applications in materials science we refer to the general reviews of Mueller et al. \cite{mueller2016machine}, Wagner et al. \cite{wagner2016theory}, Dimiduk et al. \cite{dimiduk2018perspectives} or  Wei et al. \cite{wei2019machine}.
Examples for successful applications of ML techniques in materials science are, for example, to represent inorganic materials \cite{seko2017representation, schutt2014represent, isayev2015materials}, predict fundamental  properties \cite{medasani2016predicting, de2016statistical, legrain2017chemical}, create  atomic  potentials \cite{li2015molecular}, identify functional candidates \cite{li2017high,ma2015machine, mannodi2017mining}, analyze complex reaction networks \cite{ulissi2017address}, or guide experimental design \cite{raccuglia2016machine,kim2017virtual}, high‐throughput  phase  diagram and  crystal  structure  determination \cite{graser2018machine}.

\subsubsection*{Open problem - interpretability}
However, one of the major criticisms of ML algorithms in science is the lack of novel understanding and knowledge arising from their use. This is mostly because more complex ML algorithms are often treated as black boxes.
Those  machine-built models are hard to understand for humans\cite{schmidt2019recent}. 

For a better acceptance of ML models, data scientists aim to establish clear causal relations between materials structure defined broadly across length
scales and properties. Especially scientific models have further constraints such as a minimal number of parameters and adherence to physical laws. It is the obligation of the data scientist to translate the results of their work into knowledge other 
scientists can use in aiding for example materials discovery or deployment\cite{wagner2016theory}. Useful techniques for finding simple, reduced and interpretable models are for example Principal Component Analysis (PCA) \cite{wold1987principal}, Cross-validation and regularization and a thoughtful choice of model.

PCA is a powerful technique for data dimensionality reduction. Large  datasets  are  increasingly  widespread. In order to interpret such datasets, 
PCA can be applied to drastically reduce their dimensionality in an interpretable way, such that most of the information in  the  data  is  preserved \cite{jolliffe2016principal}. PCA extracts the orthogonal directions with the greatest variance from a dataset, the resulting principal components being linear combinations of the original variables. However, principal components are not necessarily simple to interpret physically but as the extracted features are linear combinations of the original variables they  can still be intuitively explained. Moreover, it allows a very straight forward data visualization through data projection onto the main extracted components\cite{vellido2012making}. However, PCA might be the wrong choice if features are not covariant.

Another way to achieve interpretable ML models is intelligent feature selection for dimension reduction  and thus easier interpretability. Regularization of a model entails adding a tunable penalty on
model parameter size to the cost function being minimized leading to a 
reduced feature space \cite{wagner2016theory}.

Furthermore, the choice of ML model has an immediate impact on its explainability. Regressions lead to coefficients whose size gives information about the relative size effect of modifying an input on the output. Decision trees (DT) are set up like flow charts and therefore easy to read. More complex models such as Artificial Neural Networks (ANN) are missing a clear explanation of the machine’s “thinking” due to complex
node interactions. But methods such as feature visualization \cite{erhan2009visualizing,simonyan2013deep} or attribution \cite{fong2017interpretable,zeiler2014visualizing, sundararajan2017axiomatic} exist, which allow a better understanding and interpretability of black box models. However, sometimes it might be reasonable to trade model accuracy for better explainability. 


\subsubsection*{Open problem - small data}
Contrary to the curse of dimensionality, there is often the problem of small data in material characterization, since the  experimental or simulative generation of data is complex and expensive. Also the availability of the testing units limits data generation. Special test environments and material properties that are difficult to implement in the laboratory are often required, such as the creation of a corroded workpiece. 

Often models are fitted to extremely small training sets which does not play to the strength of ML and will not allow the replication of the success ML methods had in other fields. It is of course possible to use ML methods as a simple fitting procedure for small low-dimensional datasets  \cite{schmidt2019recent}.

A few approaches are known to tackle this problem. For example, a ML model can be constructed by restricting the configurational space of materials, such as predicting the band gaps of selected families of semiconductors with fixed composition or crystalline structure instead of modeling compounds spanning a wide chemical space\cite{zhang2018strategy,dey2014informatics,pilania2016machine}. Another approach by Zhang et al. \cite{zhang2018strategy} proposes to incorporate the crude estimation of the  property in the feature space to establish ML models using small sized materials data, which increases the accuracy of prediction without the cost of higher degree of freedom.

Another approach for insufficient training data  is the additional integration of prior knowledge into the training process, which leads to the notion of informed ML \cite{von2020informed}, or more specifically physics-informed ML \cite{raissi2017physics, zhu2019physics,zhang2020physics}. Domain knowledge is often given as a set of additional constraints \cite{struyf2007clustering}. The integration of additional domain knowledge generates a new hybrid formulation of the ML problem which then ideally leads to physically meaningful and significantly more accurate interpretations of the data \cite{ermon2015pattern}.
Besides adding constraints, expert knowledge can be incorporated in different ways. Up to recently, it was mostly limited to labeling data for supervised learning and setting prior probabilities in Bayesian networks \cite{macinnes2009visual}. However, in semi-supervised clustering applications, user guidance can be given by partial labeling information, which can be incorporated using hard constraints \cite{chang2007guiding}.
The domain knowledge can also be integrated  into the process of building a ML model by the application of data visualization which often improves  the accuracy of the resulting model \cite{macinnes2009visual}. Another approach is the monotonization of ML functions based on known physical relationships \cite{dette2006simple}.

In general, ML systems are rarely viewed in the context of small data, where an insufficient data size for the training model compromises the learning success. The bottleneck of the database size especially limits applications, in which the construction of a database via experiments is time consuming and costly\cite{shen2019physical}. Thus, the recent development of materials databases might be helpful in tackling the small data problem.

\subsubsection*{Existing databases}
With the launch of the Materials Genome Initiative (MGI) \cite{mgi} in 2011 and the coming of the big data era, a large effort has been made in the materials science community to collect extensive datasets of materials properties and to provide materials engineers with ready access to the properties of known materials. Existing databases are the materials project\cite{mp,jain2013commentary}, the inorganic crystal structure database \cite{icsd,international1987crystallographic}, the Materials genome initiative \cite{mgi,white2012materials} ,the NOMAD archive \cite{nr,draxl2018nomad}, the Topological Materials Database \cite{topo_db}, Supercon \cite{supercon} or the National Institute of Materials Science 2011 (NIMS) \cite{nims} with many databases of material properties of metal alloys, or National Institute of Standards and Technology (NIST) \cite{nist} with databases of properties for material classes such as structural ceramics, oxide glasses, superconductors  \cite{saal2013materials,kirklin2015open,groom2016cambridge,puchala2016materials,Zakutayev2018open,ward2018strategies}. A more comprehensive list of material databases can be found in Correa-Baena et al. \cite{correa2018accelerating}.

Traditionally, negative results are often discarded and left unpublished.  However,  negative data are  often  just  as important for ML algorithms as positive results in order to prevent bias in the data. In some disciplines with a longer tradition of data-based research (like chemistry), such databases already exist. In a similar vein, data that emerges as a side product but are not essential for a publication are often left unpublished producing so-called publication bias \cite{mlinaric2017dealing}.This eventually results in a waste of resources because other researchers then have to repeat the work in order to produce a balanced dataset for  ML applications \cite{liu2017materials}.

However, few of these available databases are ready to use with informatics techniques because they lack the uniform data formats or application programming interfaces required for informatics software \cite{ward2018strategies}.

\subsubsection*{Materials informatics}
The databases mentioned above  contain information on numerous properties of known materials and are essential for the success of materials informatics \cite{liu2017materials}.
For more general information, the reader is referred to \cite{song2004preliminary}, which gives an introducing review on materials informatics, that aptly describes the concept 
of big data in materials science. Another publication on materials informatics is, for example, \cite{wei2006materials} which introduces four main research areas in materials informatics: standardization of representation and exchange of material data; organization,  management,  retrieval,  filtration and correlation of material data; material graphics; 
and data mining and  knowledge discovery of material data. 
Another review on materials informatics mostly focuses on atomic-scale modeling \cite{ward2018strategies}. But it also promotes the idea to expand materials databases to make more data easily accessible to informatics. The widespread use of such data requires the digitalization and structuring of materials data. The data must also be easily sharable and acessible. Services that provide software interfaces to allow for automated data querying, processing, and access are evolving, for example, the Materials Data Facility \cite{blaiszik2016materials} and Citrination \cite{o2016materials,gaultois2016perspective}. Wagner et al. \cite{wagner2016theory} propose a workflow for a materials informatics problem focussing on (1) the assembly of primary features, (2) the construction of an exporatory model, (3) refinement of the model to satisfactory accuracy and (4) final training and deployment. By proceeding in an iterative fashion upwards in complexity, the final model will be as simple as possible improving its  explainability and interpretability. The shorter, but more philosophical report of Rajan et al. \cite{rajan2005materials} focuses on the role of materials informatics that allows one to survey complex, 
multiscale information in a high-throughput, statistically robust, and yet physically meaningful manner. 

\subsection{Machine learning for metallic material characterization} 
Mechanical material properties are characteristics to be precisely
predicted and controlled as they are strongly linked to and highly
affected by process parameters and resulting microstructures \cite{bock2019review}.
The
basic idea of using ML methods for material property
prediction is to analyze and map the relationships (nonlinear in
most cases) between the properties of a material and their related
characteristics by extracting knowledge from existing experimental or simulated data \cite{liu2017materials}.
Mechanical behavior in simulations is often described by
means of constitutive equations \cite{bock2019review}.\\
\\
 Research on the macroscopic performance of materials mainly
focuses on the structure-activity relationship between the macroscopic
(e.g., mechanical and physical) properties of a material and
its microstructure \cite{liu2017materials}. Many material parameters can be estimated  to  within  an  order of  magnitude using elementary physical ideas.  Whenever these parameters cannot be reliably estimated as such, ML approaches can be helpful which then require experimental or simulative data \cite{lucas2014connecting}.

Experimental testing methods which can be used on metals help understand materials and their properties better. Typical destructive  tests are bend test, impact test, hardness test, tensile test, fatigue test, corrosion resistance test or wear test, see for example \cite{meyers2008mechanical}. In the following efforts of the materials community to enhance such tests and their results with ML methods will be discussed.

\subsubsection*{Corrosion}
Corrosion detection and monitoring are essential diagnostic and prognostic means for preserving material "health" and reducing life-cycle cost of industrial infrastructures,  ships, aircraft, ground vehicles, pipelines etc.\cite{agarwala2000corrosion}. More recently, ML approaches have shown great potential to improve corrosion detection \cite{jimenez2014automatic}. They will aid a human inspector and significantly cut down on the time and cost associated with inspecting for example civil infrastructure and eliminate the need for dependence on prior knowledge and human effort in designing features \cite{atha2018evaluation}. Popular approaches are the application of ANNs for image processing-based corrosion detection.

Convolutional Neural Networks (CNN) were used for corrosion detection by investigation of images and identification of rusty parts in  the image \cite{atha2018evaluation}. The proposed CNN outperforms state-of-the-art vision-based corrosion detection approaches that are developed based on texture and color analysis using a simple multilayered perceptron network. Model input is an image of the material and region of interest, output is the classification information: corroded/not corroded for a sliding window over the image. Overfitting due to small datasets can be avoided by using pretrained networks.

Another example for the application of CNNs for corrosion detection is proposed by Bastian et al. \cite{bastian2019visual}. Similar to Atha et al. \cite{atha2018evaluation} images are used as input and classified into any of
the four classes: no corrosion, low-level corrosion, medium-level corrosion
and high-level corrosion. More investigations of similar CNN-based publications for corrosion detection are listed in \cite{bastian2019visual}.

Fang et al. \cite{fang2008hybrid} proposed a novel hybrid methodology combining genetic algorithms (GA) and support vector regression (SVR), which is capable of forecasting the atmospheric corrosion depth of metallic materials such as zinc
and steel. This hybrid approach is capable of solving nonlinear regression estimation problems in materials science. The GAs are adopted to automatically determine the optimal hyper-parameters for SVR.  The inputs for the SVR are temperature, time of wetness (TOW), exposure time, sulphur dioxide concentration and chloride concentration, respectively. The outputs are the predicted corrosion depth for zinc or steel. 

Another support vector machine (SVM) approach was applied by Hoang et al. \cite{hoang2019image} for image processing-based detection of pipe corrosion.
The image texture including statistical measurements of image colors, gray-level co-occurrence matrix, and gray-level run length is employed to extract features of the pipe surface. SVM optimized by differential flower pollination is then used to construct a decision boundary that can recognize corroded and intact pipe surfaces by blockwise classification of the original image. Also with the application of a SVR and based on a much smaller database (trained on only 46 samples), the corrosion rate of 3C steel in different environments was predicted based on five different seawater environment factors, including temperature, dissolved oxygen, salinity, pH-value and oxidation–reduction potential. The prediction error was very small \cite{wen2009corrosion}.

Jimenez et al. \cite{jimenez2014automatic} compare various ML approaches (ANNs,  SVMs,  classification  tree  (CT)  and  k-nearest  neighbour  (kNN)) for automatic pitting corrosion detection in 316L stainless steel.
Model input are environmental variables such as  chloride  concentration,  pH  and  temperature while the output is an information about the material being corroded or not. The models based  on  ANNs   and  SVM  with  linear  kernel  were demonstrated  to  be  a  valuable  tool  to  be  applied  for  this  purpose.   The classification  performance for ANN and SVM is much better  compared  to  kNN and  CT  models for  this  application. The  principal  advantage  compared  with  the traditional  techniques  is  that  it  is  not  necessary  to  apply  a surface analysis  technique  to  study  corrosion  behaviour  of  the  material.

\subsubsection*{Fatigue}
Fatigue as the weakening of a material caused by cyclic loading that results in progressive and localised structural damage and the growth of cracks can also be predicted with the incorporation of ML methods. The prediction of fatigue in welded structures for a wide range of structural materials by multi-scale FEM and ML was proposed by Shiraiwa et al. \cite{shiraiwa2017fatigue}. Two ML algorithms are applied: one is deterministic ML based on the traditional methods, and the other is model-based ML. Deterministic ML such as Multivariate Linear Regression (MLR) and ANNs use chemical composition, processing parameters (reduction ratio, heat treatment), inclusion sizes, and fatigue strength as input features to accurately predict fatigue strength.
In  the  model-based  ML,  microstructures and stress-strain curves in 40 low carbon steels with different chemical compositions and heat treatment conditions were prepared to create the learning dataset which was used to train an ANN. This approach allows for incorporation of prior knowledge of structure and property, and it can account for uncertainty such as scattering of fatigue life.

Another  attempt  to  identify  novel  connections  between  fatigue  properties and  a  variety  of  material  parameters with the help of ML algorithms has been made by Agrawal et al.\cite{agrawal2014exploration} for the prediction of fatigue strength of steel from composition and processing parameters such as chemical composition, upstream processing details, heat treatment conditions, mechanical properties. Various ML methods such as basic regression, DTs, SVM, ANN  were used. Most success showed ensemble methods and individualized methods for different types of materials.

Machine  learning  techniques  have also  been  utilized  to  predict  material fatigue life for P91 steel base metal based on the hold time in fatigue tests \cite{zhang2015using}. A combined approach of GA and SVM is used to predict the fatigue life with high accuracy.

Abdalla et al. \cite{abdalla2011modeling} use an ANN radial basis function model,
taking the maximum tensile strain and pressure ratio as
input, and put forward the model of fatigue life of steel
reinforcing bars.

\subsubsection*{Creep}
Creep is a type of metal deformation that occurs at stresses below the yield strength of a metal, generally at elevated temperatures. 
Creep rupture is becoming increasingly one of the most important problems affecting behavior and performance of power production systems operating in high temperature environments and potentially under irradiation as is the case of nuclear reactors. Creep rupture forecasting and estimation of the useful life is required to avoid unanticipated component failure and cost ineffective operation \cite{chatzidakis2014creep}. 
The  material  behaviour  is  influenced  by  the  multidimensional  interdependencies  between  the  individual  elements  of  the  chemical  composition,  the  heat  treatment  parameters, product form, tensile properties and microstructure, which are difficult to describe  using  simple  analytical  methods.  Modeling  with  ML  techniques  therefore  seems  to  be  an  interesting  alternative.  Moreover,  the  application of ML  takes  away the requirement for long and expensive experiments \cite{frolova2011representation}. 

For the design of materials, creep is considered an important material property. However, quite often such designs only focus on one objective (e.g. creep) without considering the comprehensive design of multi-property \cite{wang2019design}. 
For the investigation of  creep rupture life and rupture strength of austenitic stainless steels\cite{sourmail2002neural} once again ANNs are popular models. For the prediction of  creep rupture life and the creep rupture stress for a given stress, the training database for the input parameters contains test conditions (stress and temperature), chemical composition, solution treatment temperature  and time (the latter being available in a very limited number of  cases), nature of  the quench following, grain size, and logarithm of  ruptured life for a broad variety of stainless steels. 

Chatzidakis et al. \cite{chatzidakis2014creep} employ and compare General Regression Neural Networks, ANNs and Gaussian Processes to capture the underlying trends and provide creep rupture forecasting. Input parameters are  experimental creep rupture data. However, the overall performance of the developed models was insufficient.

In the study of Shin et al. \cite{shin2019modern}, the five different ML models random forest (RF), linear regression (LR), kNN, kernel ridge (KR), Bayesian ridge (BR) are applied for the prediction of Larson-Miller parameters which represent the creep behavior. From a wide range of available features (466), relevant ones are selected with optimization approaches and different set-ups of features and models evaluated. Highest accuracy was obtained by RF for a varying number of top ranking features between 5 and 21.

For the prediction of rupture  and  creep  rupture  stress of 9$\%$Cr steels, a Multilayer Perceptron Neural Network was applied with the input parameters chemical composition, heat treatment information, geometrical form of the investigated components \cite{frolova2011representation}.


\subsubsection*{Flow behaviour and work hardening}
In sheet metal forming operations, mechanical properties of the sheet material such as flow stress or stress–strain curves greatly influence metal flow and product quality \cite{gutscher2004determination}. The flow stress can be determined by tensile tests which provide uniaxial information about stress–strain  behavior.
Hardness tests measure the resistance of material to an indenter, and hardness correlates well with flow stress. Compression testing reveals flow stress but is complicated by friction and buckling. Bend testing and torsion testing provide a good measure of flow stress and fracture resistance but are complicated by radial stress and strain gradients\cite{wright2016wire}. 

One of the main advantages of applying ML approaches  is that it is not necessary to postulate a mathematical model at first, which is quite difficult because of the nonlinearities in the response of the deformation behaviors of
the materials under elevated temperatures and strain rates and the factors affecting the flow stress. Lin et al. \cite{lin2008application} propose a feed forward back propagation ANN model to predict the constitutive flow behaviors of 42CrMo steel during hot deformation, and investigate the general nature of the influence of strain, strain rate and temperature on the compressive deformation characteristics of 42CrMo steel. The capability of the developed ANN model to predict the flow
stress level, the strain hardening and flow softening stages is also investigated. The inputs of the ANN are deformation temperature, log strain rate and strain whereas flow stress level, the strain hardening and flow softening stages are the output. Low absolute relative errors for training and test data promise a good generalizing model.

Gupta et al.\cite{gupta2013development} predict flow stress based on a feed forward ANN trained with the back propagation algorithm. Input features were strain, strain rate and temperature. The implemented ANN produced more accurate results than conventional mathematical  models  such  as  Johnson  Cook,  modified Zerilli-Armstrong  and  modified Arrhenius. 
SVR based flow stress prediction for austenitic stainless steel 304 with strain,  strain  rate  and  temperature as inputs and the flow stress as output was proposed by Desu et al. \cite{desu2014support} and provides more accurate results than the conventional mathematical  models.

The identification of work hardening properties of steel and an aluminum alloy from indentation tests is described by Meng et al. \cite{meng2015identification}. The authors  propose a material parameter identification protocol based
only on the imprint shape of the indentation test using orthogonal decomposition and manifold learning for the prediction of the strain hardening exponent and yield stress.

 ANN modeling for anisotropic mechanical properties and work hardening behavior of Inconel 718 alloy at elevated temperatures was proposed by Mahalle et al. \cite{mahalle2019neural}. Strain and temperature are used to successfully predict material properties such as ultimate strength, yield strength, elongation and strain hardening coefficient.

\subsubsection*{Tensile properties}
 Tensile properties indicate how the material will react to forces being applied in tension. Determining the tensile properties is crucial because it provides information about the modulus of elasticity, elastic limit, elongation, proportional limit, reduction in area, tensile strength, yield point, yield strength, and other tensile properties \cite{rahman2019tensile} which then define the state of the material, its longevity or its ability to perform in an application. Thus, the accurate prediction of tensile properties has great importance for the service life assessment of structural materials \cite{wang2020tensile}.
 
 The ultimate tensile strength (UTS) of iron castings which gives information about the capacity of a metal to resist deformation when subject to a certain load was predicted based on a variety of input features such as composition, size of casting, cooling speed or thermal treatment (25 variables) to gain information about 
the mechanical properties of a foundry and thus predict foundry defects \cite{santos2009machine}. Estimating the value of UTS is one of the hardest issues in foundry production, due to many different circumstances and variables that are involved in the casting process. Bayesian networks, kNN and ANNs were used for classification of the UTS. All of the investigated approaches perform well, but ANNs outperformed the other classifiers. 
Similar to the aforementioned study, Sterjovski et al. \cite{sterjovski2005artificial} also used ANNs (of back propagation type ANN) to predict mechanical properties of steel, such as 
the impact toughness of quenched and tempered pressure vessel steel exposed to multiple postweld heat treatment cycles; the hardness of the simulated heat affected zone in pipeline and tap fitting steels after in-service welding; and the hot ductility and UTS of various microalloyed steels over the temperature range for strand or slab straightening in the continuous casting process. Input parameters were composition, cooling rate, temperature and thickness. It was shown that ANNs could successfully predict all mechanical properties investigated.
Another example for the application of ANNs  is proposed by Sankar et al.\cite{sankar2009applying} who predict elongation, self-tempering temperature and yield strength for  reinforcement steel bars subjected to thermomechanical treatment based on two input parameters (bar diameter and quenching duration). The numerical results of the ANN are compared with experimental results and are found to be in good agreement.

Pruning and predator prey algorithms were applied by Datta et al.\cite{datta2007designing} which were able to extract more knowledge from the input data, than typically possible with conventional ANN analysis. Alloy composition and the  thermo-mechanical controlled processing parameters,  deformation  in  different temperature  zones,  finish  rolling  temperature  and cooling  rate  of  high  strength steels have  been  taken  as  input  parameters, whereas  UTS,  yield strength  and percentage  elongation were predicted. It was shown  that in this type of steel the yield strength depends
mostly on the solid solution hardening and the microstructural
constituents while UTS is more influenced by the precipitation hardening, but all these
strengthening mechanisms have a negative effect on the
ductility of the steel. 

Pattanayak et al. \cite{pattanayak2015computational} investigated the role of the composition and processing parameters on the mechanical properties of API grade microalloyed pipeline steel, in respect to its strength, impact toughness and ductility. ANN models, capable of prediction and diagnosis in non-linear and complex systems, are used to obtain the relationship of composition and processing parameters with said mechanical properties. Then the models are used as objective functions for the multi-objective GAs for evolving the tradeoffs between the conflicting objectives of achieving improved strength, ductility and impact toughness. The Pareto optimal solutions were analyzed successfully to study the role of various parameters for designing pipeline steel with such improved performance.

For the ANN-based prediction of  yield  strength,  UTS,  ductility (elongation, and  reduction  of  area) of  ferritic  steel  weld  metals  appropriate  for  the welding of high strength low alloy steels, Metzbower et al.\cite{metzbower2001neural} use the  chemical  composition  of  as  deposited  weld  beads, and the cooling rate. The various established ANN models are found to work well once combined to an ensemble of best models. The ensemble produced a more  reliable  prediction  than  an individual  model and reproduced known metallurgical trends well. Poudel et al. \cite{poudel2013selective} also compare different ML models and ensembles for the  prediction of UTS, yield strength, elongation of steel bars instead of performing tensile tests. Input parameters were process parameters from the rebar manufacturing process such as for example material composition, temperatures, rod diameter, rod speed and cooling rate. Investigated classification algorithms were among others MLR, Principal Components Regression, Partial Least Squares Regression, ANNs and Locally Weighted Regression. The main conclusion from this research is that any single model cannot efficiently describe the complicated relationship
of the input-output space of the rebar manufacturing process. Ensemble methods, however, as combination of ML approaches used together in conjunction with some model selection or model weighting techniques lead to robust prediction systems.

Applications of traditional ML algorithms for the prediction of tensile properties were proposed by \cite{shigemori2007optimum,shigemori2011optimum,swaddiwudhipong2005material,wang2019design,wang2020tensile}. Shigemori et al.\cite{shigemori2007optimum,shigemori2011optimum} reported
about the successful application of Locally Weighted Regression (LWR) in predicting the tensile strength for a certain  type  of  steel  product  which  is  produced  by  hot  rolling. As input, 18 items were selected  from
chemical composition, heating, rolling, and cooling temperature.
These variables have a clear physical causal relationship with the
output variable. 
Furthermore, least squares SVMs are suitable approaches for the prediction of the elastic modulus and yield stress of materials. In \cite{swaddiwudhipong2005material} FEM-simulated load-indentation curves of 
Al6061 and Al7075 are investigated for the determination of these material parameters based on a training set of large strain-large deformation FEM for the simulation of indentation tests. Characteristic features are extracted from the load-indentation curves and used in the ML model. The proposed  least squares SVM model is capable of predicting reasonably accurately the elastic modulus and yield stress of materials based on the load-indentation curves of dual conical indenters with different half-angles.
For material design, a RF model in combination with an optimization algorithm was found to relate yield strength, impact toughness and total elongation with material composition information and treatment parameters for the production of RAFM steels \cite{wang2019design,wang2020tensile}. For yield strength, highly correlated features were tempering temperature and C content  and tempering time
and Cr content for elongation. The accuracy and  generalization ability of the RF was  acceptable ($R^2>85\%$).

An interesting study different from most approaches using  material composition information as input features for material parameter prediction was proposed by Fragassa et al. \cite{fragassa2019predicting} for the prediction of the tensile behaviour of cast alloys such as yield
strength, ultimate strength, ultimate strain and Young’s modulus, by a pattern recognition analysis  on experimental data and the application of RF, ANN, kNN.
All Information is directly taken from micrographs. For the prediction of UTS and yield strength ANNs show the best results.

Another application of ANNs is the interpretation of acoustic emission data for failure prediction. Christopher et al. \cite{christopher2018neural} propose the prediction of  the  ultimate  strength  of  aluminum/silicon  carbide  (Al/SiC)  composites  by  using  acoustic  emission parameters  through ANN analysis.
This approach was earlier pursued for the prediction of the ultimate strength of unidirectional T-300/914 tensile specimens using acoustic emission response and an ANN back propagation algorithm \cite{sasikumar2008artificial}.

\vspace{0.3cm}
To prevent surface cracks on cast steel, its hot ductility must be monitored. Experimental investigations are difficult to execute. Thus, ML models are proposed for the prediction of hot ductility. For example, a MLR \cite{liu2012control} predicted hot ductility and
grouped 12 chemical elements that had similar experimental effects on ductility. The cooling condition used here is different from actual continuous-casting conditions, so the model is difficult to apply in commercial setups. A back-propagation ANN  method has been
used \cite{sterjovski2005artificial} to predict hot ductility for various microalloyed steels over the temperature range for strand or slab straightening in the continuous casting process. However, the recorded data are limited and do not lead to a generalizing model. Additionally, the NN used
only one hidden layer, so it has a limited ability to describe
the complex relationship between input and output.
Also an ANN model was used to predict high-temperature ductility of various steel grades from their composition and thermal history (described  by five experimental variables) \cite{kwon2019prediction}. The developed model can predict ductility for a wider composition range and thermal
history than previous studies have achieved. Therefore, it
can be used effectively in commercial production.


\subsection{Machine learning for the small punch test}

As mentioned above, the estimation of tensile properties of materials such as elastic modulus, yield strength or strain hardening exponent is considered to be of fundamental importance. Conventional tests are  destructive in nature and require reasonable specimen cross-section and volume. There are situations where limited volume of material is available for property assessment, like in material development, failure analysis and remaining life assessment of in-service components, materials for pressure vessels, turbines, thermal power plants or chemical processing industries \cite{husain2017small,altstadt2018estimation}. In these situations, test techniques using small volume specimens become more attractive. Small specimen test techniques have been established as a reliable alternative to the traditional tensile test, as the results of these test techniques are in good agreement with the tensile test and are reproducible when tested under controlled conditions \cite{arunkumar2016estimation}. 

The extraction of mechanical properties of in-service materials through such small specimen test techniques has become more popular in recent years. Of the available small specimen test techniques, SPT methods have proven to be promising \cite{wang2017determination, manahan1981development, manahan1982development, arunkumar2016estimation}. These test methods are basically non-destructive in nature and are proficient enough to extract the flow properties of the materials using small volume specimen. 
After more than 40 years’ development, there are some standards in SPTs, such as ASTM-F2183 \cite{astm}, GB/T-29459 \cite{gbt}, and CEN CWA-15627\cite{cwa}.

A prerequisite for using this test is to establish correlations between SPT and conventional tests \cite{arunkumar2020overview, janvca2016small}, such as relations between the tensile test and the Erichsen cupping test for the determination of the critical damage value curves, the initiation time and location of fracture \cite{kim2014comparative}. Other examples are the correlation between the SPT and tensile tests for the estimation of the tensile strength \cite{klevtsov2008using} or the construction of stress–strain curves only from indentation tests \cite{n2013mechanical}. Here the link between mean pressure and the indent diameter, obtained from indentation tests, to the stress and strain determined from a tensile test was obtained by the empirical relation of Tabor \cite{tabor1951hardness}. 

Exploiting force-displacement curves (FDC) from spherical indentation simplifies the identification of mechanical properties. Parameters such as hardness, deformation mode, yield stress, Young's modulus of the indented material can also be extracted out of the imprint shape, which is a valid alternative for using the indentation curve \cite{meng2017objective}. For example, Milivcka et al. found linear correlations between the SPT maximum force and the tensile strength of a 9$\%$ Chromium steel for creep resistant high temperature applications \cite{milivcka2006small}.

Detailed  analyses  of  stress  and  strain  in  the  SPT  disc  have been  performed  by  means  of  analytical  elastic-plastic  modelling, see for example \cite{chakrabarty1970theory,byun2001characterization} and  by  FEM to  underpin  the  empirical  correlations \cite{campitelli2004assessment, abendroth2006identification, simonovski2017small}.
FEM simulations are mainly used to generate FDCs based on an assumed material constitutive law which are then compared with experimental FDCs. This approach is called inverse analysis \cite{linse2008usage,lu2020extraction}. For more information, see for example \cite{penuelas2011inverse,wei2019machine}. However, the solution to the inverse identification problem is non-unique. Furthermore, there are other problems such as insufficient accuracy, indentation frame/machine compliance, noisy input data, difficulty in determining the exact starting point of the load (force)-displacement curve, bending of the specimen in thin-sheet indentation and FE mesh dependence \cite{meng2017objective}.

Various mechanical properties that can be extracted through SPT, are for example
fracture mode \cite{kameda1986kinetic,misawa1987small,mcnaney1991application}, 
 yield  stress \cite{kameda1992small,campitelli2004assessment,linse2014quantification},  
UTS \cite{mao1987development,hurst2012we,wang2017determination} (based on hydraulic bulge tests which are similar to SPT but high-pressure hydraulic oil is used instead of punch to cause specimen deformation);
 tensile strength \cite{milivcka2006small},
 Young’s  modulus \cite{oliver1992improved,field1993simple,taljat1997analysis, huber1997determination},
 fracture  toughness \cite{abendroth2006identification,linse2008usage,konopik2013determination,wang2008small,bulloch1998toughness,mao1991small,misawa1989fracture},
 creep  properties \cite{dymavcek2009creep,gulccimen2013determination} or 
 elastic plastic properties \cite{sainte2002small, fleury1998small}.
 
These examples show that the SPT is a  widely used approach for the determination of various material parameters from a small amount of material. However, the approach also has a few disadvantages, which have to be kept in mind. To begin with, the selected sample size may not represent the bulk material. The sample size effect has an important influence on
the mechanism of the fracture of SPT, because often the thickness of the
SPT sample is only 5–8 times of the mean grain size.
The changes in grain-boundary distribution and the grain orientation,
that resulted from the sample size effect, may affect the
mechanical behaviors of the SPT samples. These parameters may
affect the mechanism of fracture on small scale material samples \cite{song2012comparison}
In addition, SPT response is sensitive to various test parameters such as specimen shape, specimen thickness, test speed, ball diameter, clamping force and material. For more information, the reader is referred to \cite{arunkumar2020overview}, where an extensive collection of SPT configurations is listed.

\vspace{0.3cm}
In recent years, using ML approaches on SPT data has become popular.
Most approaches incorporate ANNs for the identification of material properties from SPT data. Abendroth et al. \cite{abendroth2016assessment} identify ductile damage and fracture parameters from the SPT using ANNs. FEM is used to compute the load displacement curves.  Via a systematic variation of the material parameters a data base is built up, which is used to train the ANNs. This neural network can be used to predict the load displacement curve of the SPT for a given material parameter set.  The identified material parameters are validated by independent tests on notched tensile specimens. 
A similar approach was followed by Abendroth et al. \cite{abendroth2003determination, abendroth2006identification}  earlier for the  estimation of  the   hardening and damage parameters of  high strength steels. The  combination  of FEM,  SPT data  and  ANNs  was used  to  identify  the parameters  of  the  Gurson–Tvergaard–Needleman  model  for  ductile  damage and fracture parameter prediction.

Linse et al. \cite{linse2008usage} use synthetic load displacement curves generated via FEM for a variety of material parameters. This database is then used to train ANNs, which approximate the load displacement curves of the SPT as a function of the material parameters. The identification procedure itself consists of an optimisation algorithm, minimising the difference between the measured load displacement curves and its approximation by the neural networks until the true set of material parameters is found. Prerequisite is an accurately working FEM-Model.
The approach was applied for the identification of hardening parameters and WEIBULL-parameters in the brittle and brittle–ductile transition region of two non-irradiated reactor vessel steels.

Another employment of FEM simulations of tensile, bulge, Erichsen tests is proposed by Abbassi et al. \cite{abbassi2013parameter} for the calculation of damage parameters. The Gurson–Tvergaard–Needleman (GTN) model was employed. An identification procedure based on ANNs is used to determine the material parameters of the GTN damage model. The ANN was trained by using the FEM results of the notched tensile test with varying  the damage parameters. A good efficiency for the identification of damage parameters was proven.
The Poisson ratio can also be predicted using a similar approach (combination of SPT and FEM and ANN) \cite{huber2001determination}.  
Using  FE simulations, the relation between the material parameters and the quantities characterizing  the  depth-load  response  is  calculated.  An  approximate  inverse  function represented by an ANN is derived on the basis of these data.

\vspace{0.3cm}
In contrast to all the ANN-based approaches described so far, the study of Meng et al.\cite{meng2017objective} suggests the use of manifold learning for indentation-based material characterization. However, FEM was also used there. Input parameters were features describing the imprint shape of SPTs. The manifold learning approach was able to iteratively reduce the distance between the FE-simulated and the experimental imprint shapes in order to identify the material hardening parameters. The approach was successfully shown for three different materials: AISI1095 steel and two aluminum alloys EN AW-2017F and EN AW-5754F.

Another innovative approach suggests the application of digital video processing of the forming process created by the  Erichsen cup test \cite{sangkharat2019using}.
The OpenCV  library \cite{bradski2008learning} was  used  to  develop  a  sheet-metal-forming  image  analysis  program which identifies the indent shape and position to calculate the anisotropic coefficient which showed good agreement with tensile test results. However, results were biased by the environment around the testing area.

 A  code  of  practice  on  SPT  was established in 2010 in  order  to  harmonize  the  various  test  set-ups  and  to  achieve  a  better comparability  of  the  results  of  different  labs \cite{hurst2010european}.

\section{An application of ML in material property prediction}\label{application}
\begin{center}
\begin{table}
\begin{tabular}{ lllll } 
\hline
 Material & Type & Product form & Heat treatment & Reference\\
 \hline
 P91 & f/m steel & Hot rolled pipe $\diameter$ 360 $\times$ 50 mm  & Normalization 1040–1100$^\circ$C/30min&\cite{kohlar2017gefuge}\\
 &&& tempering 730–780$^\circ$C/60min &\\
P92 & f/m steel & Hot rolled pipe $\diameter$ 219 $\times$ 22 mm  &AR: standard normalization + tempering &\cite{Houska2017}\\
&&&HT1: AR + 800$^\circ$C/2h&\\
&&&HT2: AR + 760$^\circ$C/2h&\\
&&&HT3: AR + 750$^\circ$C/2h&\\
&&&HT4: AR + 740$^\circ$C/2h&\\
Eurofer97 & f/m steel & Hot rolled plate 14.5 mm & Normalization 980$^\circ$C/27min/air cooling &\cite{Heintze2009ion,Tavassoli2004materials}\\
&&&tempering 760$^\circ$C/90min/air cooling &\\
22NiMoCr 37 & Bainitic steel & Reactor pressure vessel Biblis C & 890$^\circ$C/4 h/water quenching  & \cite{Zurbuchen2009influence}\\
&&&650$^\circ$C/7h/air cooling&\\
15Kh2MFA& Bainitic steel & Reactor pressure vessel Greifswald &Original RPV manufacturing technology & \cite{Viehrig2006application}\\
&& unit 8&&\\
 \hline
\end{tabular}
\caption{Materials overview.}
\label{table_materials}
\end{table}
\end{center}
Similar to the studies mentioned above, we want to apply ML models to extract material properties from SPT data since the SPT has long been recognised as a supportive means for the development and monitoring of structural materials.

Conventionally, the following empirical correlation has been used for the estimation of the UTS $R_m$ from FDCs \cite{campitelli2004assessment,mao1987development,dymavcek2013investigation,kumar2015evaluation, bruchhausen2016recent}

\begin{equation}
R_m=\beta_{R_m} \cdot F_m / (h_0\cdot v_m)
\label{eq2}
\end{equation}
with $F_m$ being the maximum force, $v_m$ the corresponding
punch displacement, $h_0$ the initial specimen thickness and $\beta_{R_m}$ an empirical coefficient. However, various researchers \cite{altstadt2018estimation, kumar2015evaluation} propose the determination of $R_m$ based on the force $F_i$ instead of $F_m$ for the correlation with the UTS in order to avoid a strong dependence of the correlation factor $\beta_{R_m}$ on the tensile properties of the material:
\begin{equation}
R_m=\beta_{R_m} \cdot F(v_i) / ({h_0}^2)
\label{eq5}
\end{equation}
with $\beta_{R_m}$ being geometry and material dependent. More information about this empirical approach can be found in \cite{altstadt2018estimation}.
In addition to the empirical approach (1),(2), we propose two ML models to determine the UTS $R_m$ from SPT data, shown in Figure \ref{figure_Rm}(b).

\subsection{Experiments}
\begin{figure}
   
   \begin{minipage}[b]{.49\linewidth} 
       \begin{overpic}[scale=0.49,,tics=10]
          {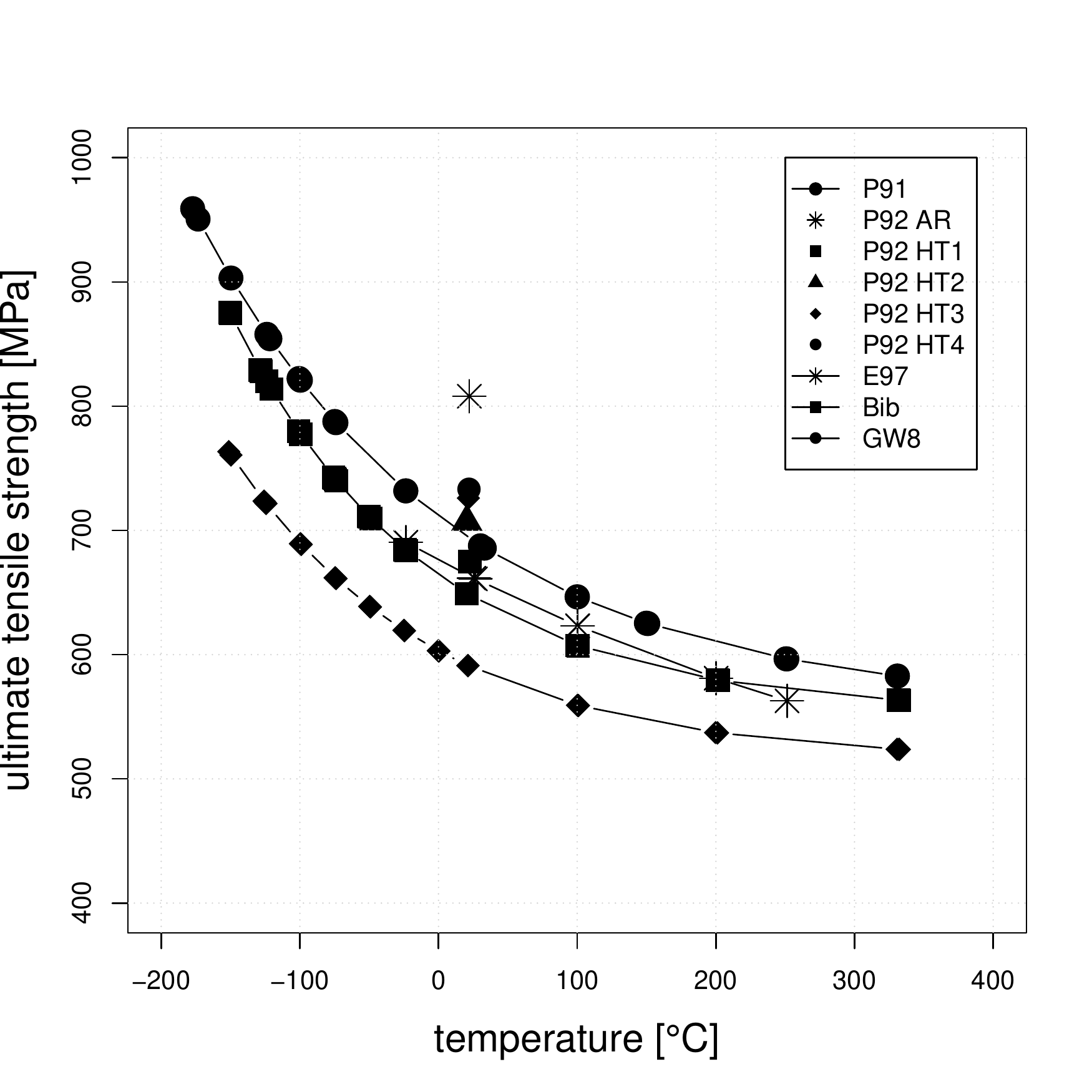}
          \put(-1,84){(a)}
      \end{overpic}
    \end{minipage}
   \hspace{.05\linewidth}
   \begin{minipage}[b]{.49\linewidth} 
      \begin{overpic}[scale=2,,tics=10]
          {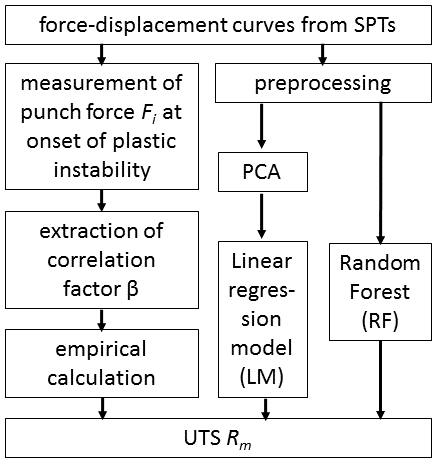}
          \put(-6,94){(b)}
      \end{overpic}
   \end{minipage}
   \caption{(a) UTS $R_m$ for the investigated materials and temperatures. (b) scheme.}
   \label{figure_Rm}
\end{figure}

\begin{figure}
   \begin{minipage}[b]{.49\linewidth} 
       \begin{overpic}[scale=0.49,,tics=10]
          {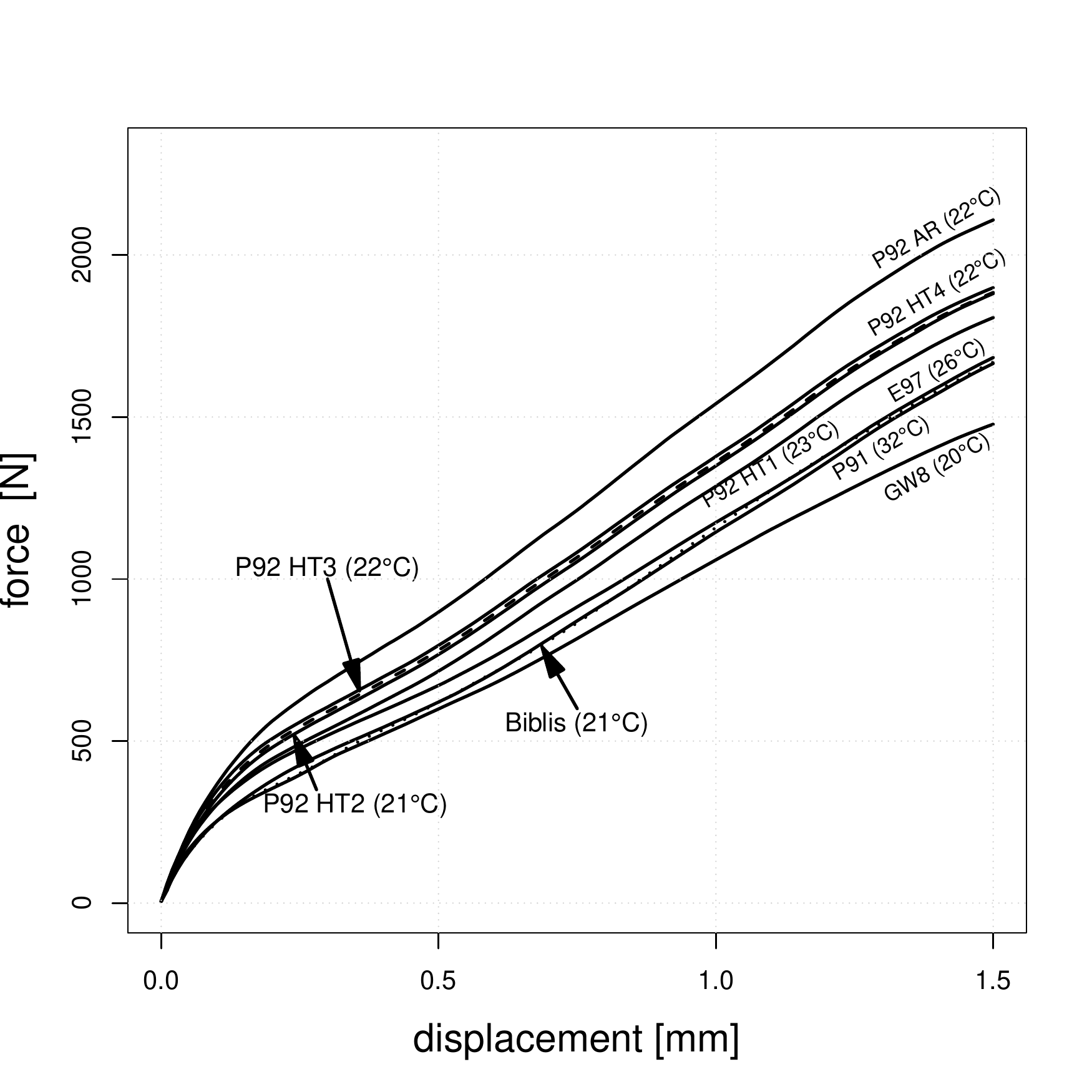}
          \put(-1,85){(a)}
      \end{overpic}
   \end{minipage}
   \begin{minipage}[b]{.49\linewidth} 
         \begin{overpic}[scale=0.49,,tics=10]
          {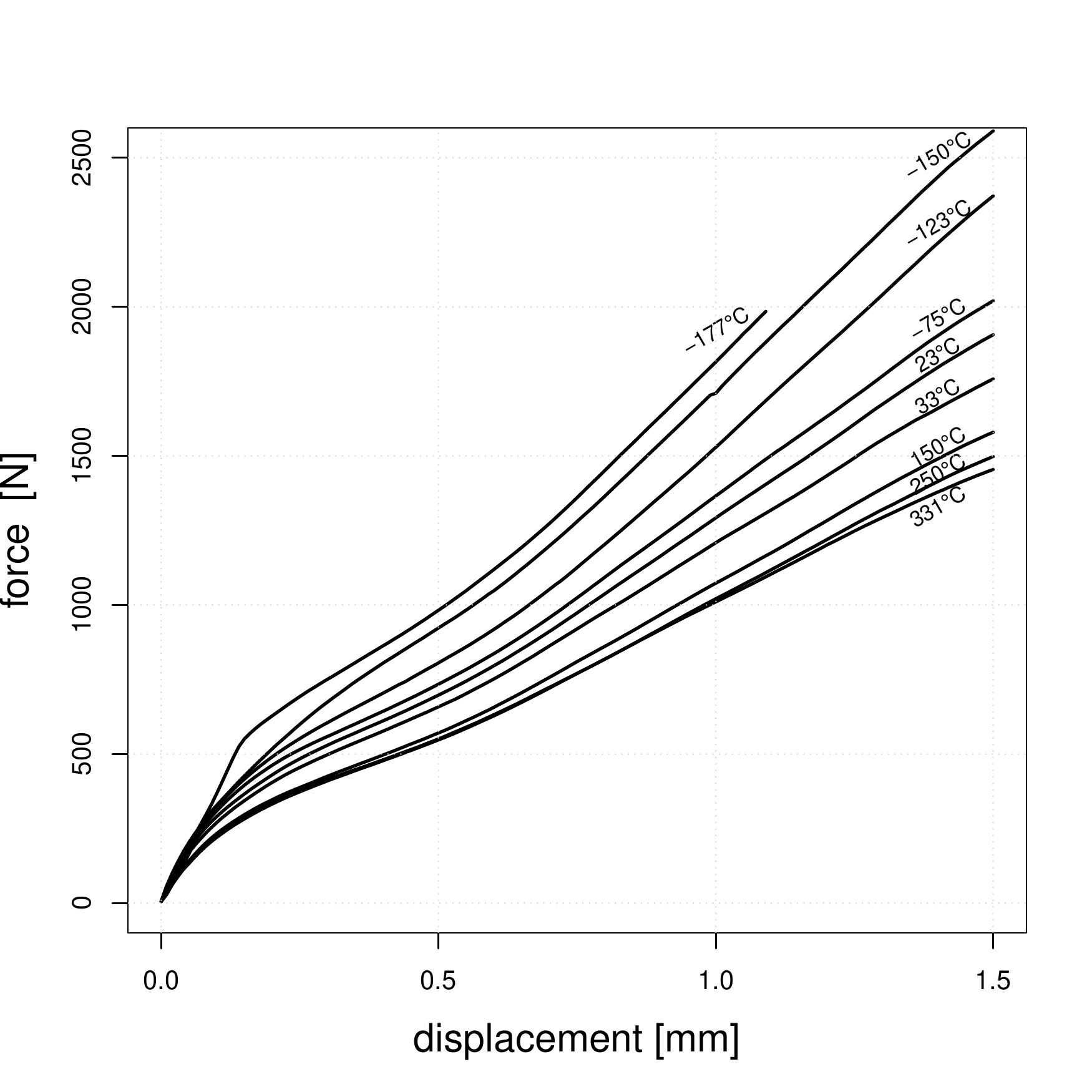}
          \put(-3,85){(b)}
      \end{overpic}
   \end{minipage}
   \caption{(a) SPT data at room temperature for materials listed in Table \ref{table_materials} \hspace{-0.3cm}, (b) SPT data for P91 at different temperatures.}
   \label{SPTdata}
\end{figure}

SPT data of three ferritic-martensitic (f/m) Cr-steels and two bainitic reactor pressure vessel steels\cite{altstadt2018estimation} were used to predict the UTS $R_m$ based on ML approaches. The UTS $R_m$ was obtained by tensile tests. A materials overview can be found in Table \ref{table_materials}, which was adopted from  Altstadt et al.  \cite{altstadt2018estimation}. The P92 was available in the, as received condition, and in four different heat treatments. The specimens were of 0.5 mm thickness. SPT were conducted with a punch diameter of 2.5 mm, a receiving hole diameter of 4 mm, edge size 0.2 mm and the edge type chamfer.
All tests were performed at a displacement rate of 0.5  mm/min. The displacement $v$ was measured by an inductive sensor with an accuracy of $\pm$ 1 $\mu$m.
The FDCs $F(v)$ of SPTs at room temperature are shown in Fig. \ref{SPTdata} (a) (mean of all available curves per material at room temperature). The $F(v)$ curves of the material P91 for different temperatures are shown in Fig. \ref{SPTdata} (b). The curve for $T = -177^\circ C$ represents a brittle failure, the other test curves represent ductile failure. By using 9 different materials and heats, and by testing selected materials at different temperatures, a wide variation of tensile properties is obtained.
Table \ref{table_tests} gives the number of conducted tests and their test temperatures for each type of material. Figure \ref{figure_Rm} shows the UTS $R_m$ for all investigated materials and temperatures.

For more detailed information on the experimental set-up the reader is referred to Altstadt et al. \cite{altstadt2018estimation}.
\begin{table}
\centering
\begin{tabular}{ lll } 
\hline
 Material & Number of tests & Test temperatures\\
 \hline
 P91&23& -177...+331\\
 P92-AR& 3& Room temperature\\
 P92-HT1 & 5 & Room temperature\\
 P92-HT2 &4& Room temperature\\
 P92-HT3 &3&Room temperature\\
 P92-HT4 &5&Room temperature\\
 Eurofer97 &8& -24...+250\\
 22NiMoCr37 & 31& -151...+332\\
 15Kh2MFA & 33&-150...+332\\
 \hline
\end{tabular}
\caption{Overview of SPTs.}
\label{table_tests}
\end{table}

\subsection{Preprocessing}
First, unified sampling points for the displacement were defined with a distance of  44 $\mu$m based on a sampling rate of roughly 1 $\mu$m in the raw data. Then each of the displacement values is defined as a feature, leading to 151 features for displacement values up to 1.5 mm. Failure occurred for displacements larger than 1.5 mm. An additional feature is the test temperature.
Now, each time series is considered to be an observation. Dimensions correspond to the number of displacement sampling points + 1 (temp).

As the number of dimensions (p) increases, the volume of the domain increases exponentially. This, in turn, requires more samples (n) from the domain to provide effective coverage of the domain for a learning algorithm. This problem was introduced earlier as the curse of dimensionality \cite{james2013introduction}. ML algorithms overcome the curse of dimensionality by making assumptions about the data and structure of the mapping function from inputs to outputs which adds a bias and leads to a loss of the generalization power of the ML model\cite{james2013introduction}. Ways to approach this problem are feature selection, projection methods or the application of regularized algorithms.

\begin{figure}
   \begin{minipage}[b]{.49\linewidth} 
       \begin{overpic}[scale=0.49,,tics=10]
          {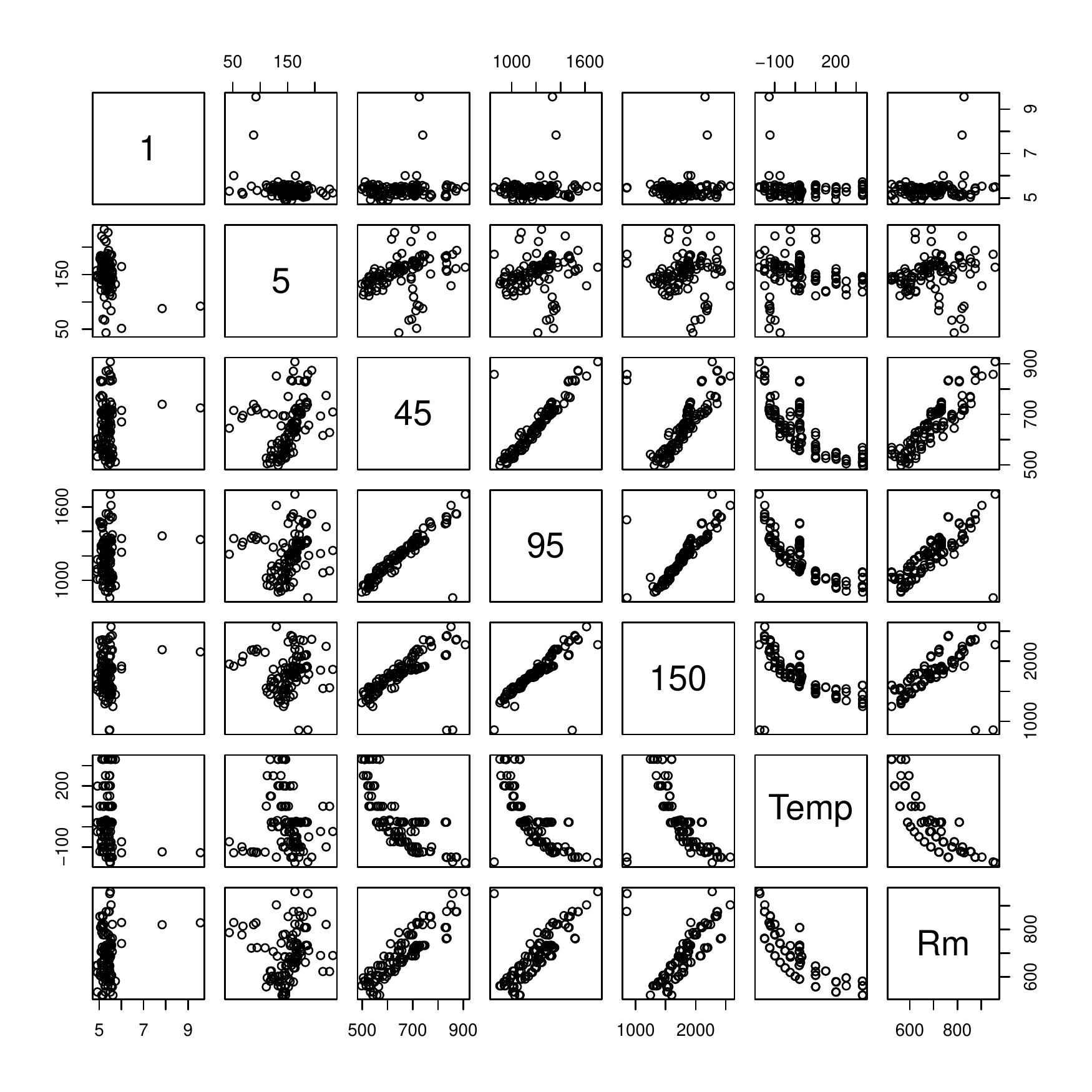}
          \put(-1,84){(a)}
      \end{overpic}
   \end{minipage}
   \hspace{.05\linewidth}
   \begin{minipage}[b]{.49\linewidth} 
       \begin{overpic}[scale=0.49,,tics=10]
          {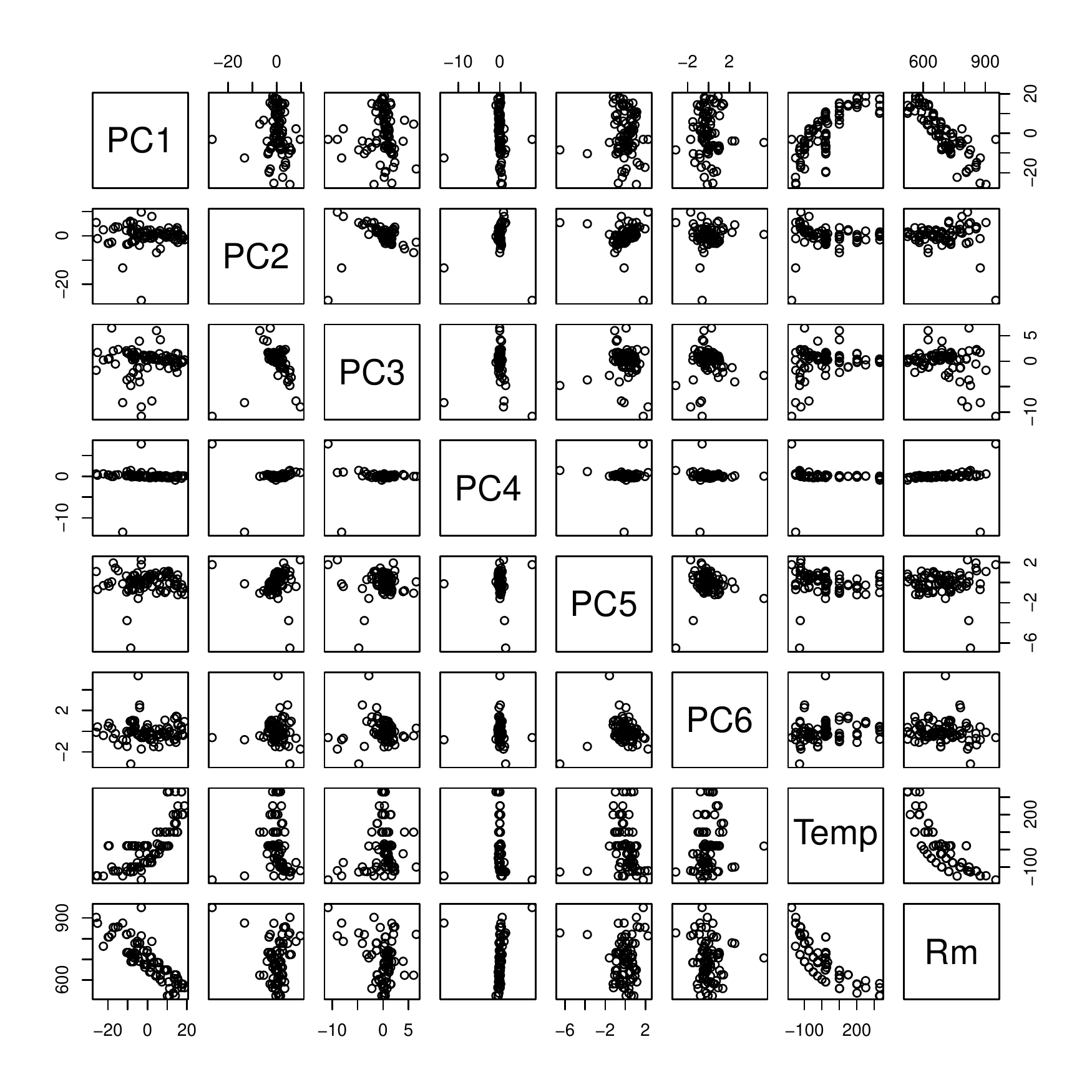}
          \put(-3,84){(b)}
      \end{overpic}
   \end{minipage}
   \caption{(a) Pairs plot of selected features, temperature and $R_m$, (b) pairs plot of first 5 principal components, temperature and $R_m$.}
   \label{preprocessing}
\end{figure}

Another problem when using FDC-data as input data for a ML algorithm is the multicollinearity of the features extracted from these curves. Exemplary correlations between features 1, 5, 45, 95, 150 and the temperature  are shown in Figure \ref{preprocessing} (a).
Subsequently, principal component analysis (PCA) is performed on the FDC-data in order to eliminate co-linearity in the features. The broad idea behind this scheme is that, in contrast to the original features, the principal components will be uncorrelated. Furthermore, one expects that a small number of principal components will explain most of the variance and therefore provide an accurate representation of the dataset \cite{schmidt2019recent}.  For the present dataset a total of 6 principal components explain 99$\%$ of the overall variability. Now, the pairs plot  of the new features in Figure \ref{preprocessing}(b) shows only horizontal or vertical lines between input features (principal components) which is typical for uncorrelated features.

\subsection{Machine learning models}
For the estimation of tensile properties such as the UTS $R_m$ from SPT data, empirical equations such as (1) or (2) can be used. The force $F_i$ can be associated with the onset of plastic instability and is therefore well suited for a correlation with the UTS $R_m$ of the uniaxial tensile test as proposed in \cite{altstadt2018estimation}.
With the number of available data points being rather small a linear regression model (LM) in combination with PCA and a RF based on the original SPT data are proposed for the prediction of $R_m$.
The root mean squared error (RMSE) was used to compare the different approaches. 

\begin{figure}
   \begin{minipage}[b]{.3\linewidth} 
  \begin{overpic}[scale=0.3,,tics=10]
          {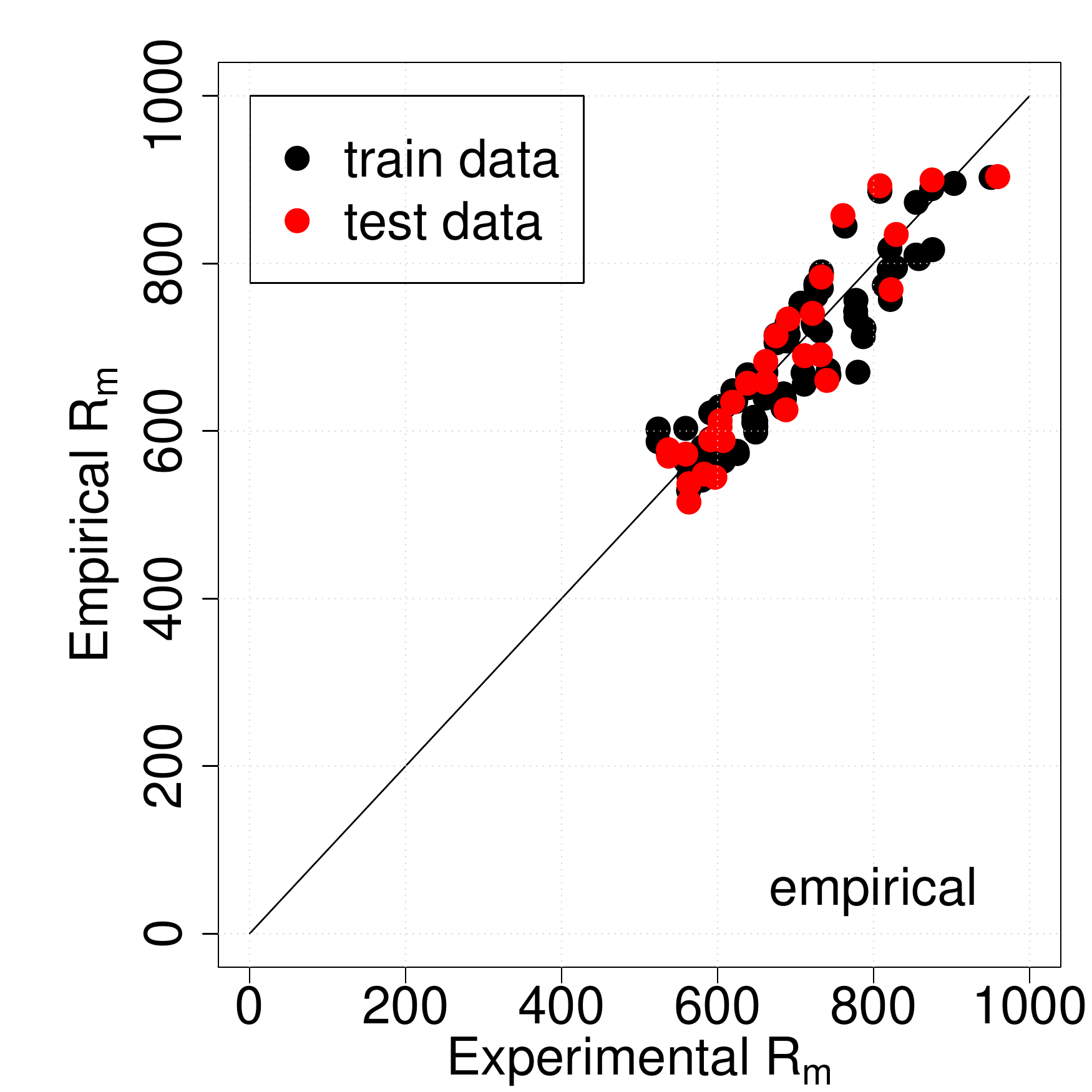}
          \put(-1,84){(a)}
      \end{overpic}
   \end{minipage}
   \hspace{.05\linewidth}
   \begin{minipage}[b]{.3\linewidth} 
   \begin{overpic}[scale=0.3,,tics=10]
          {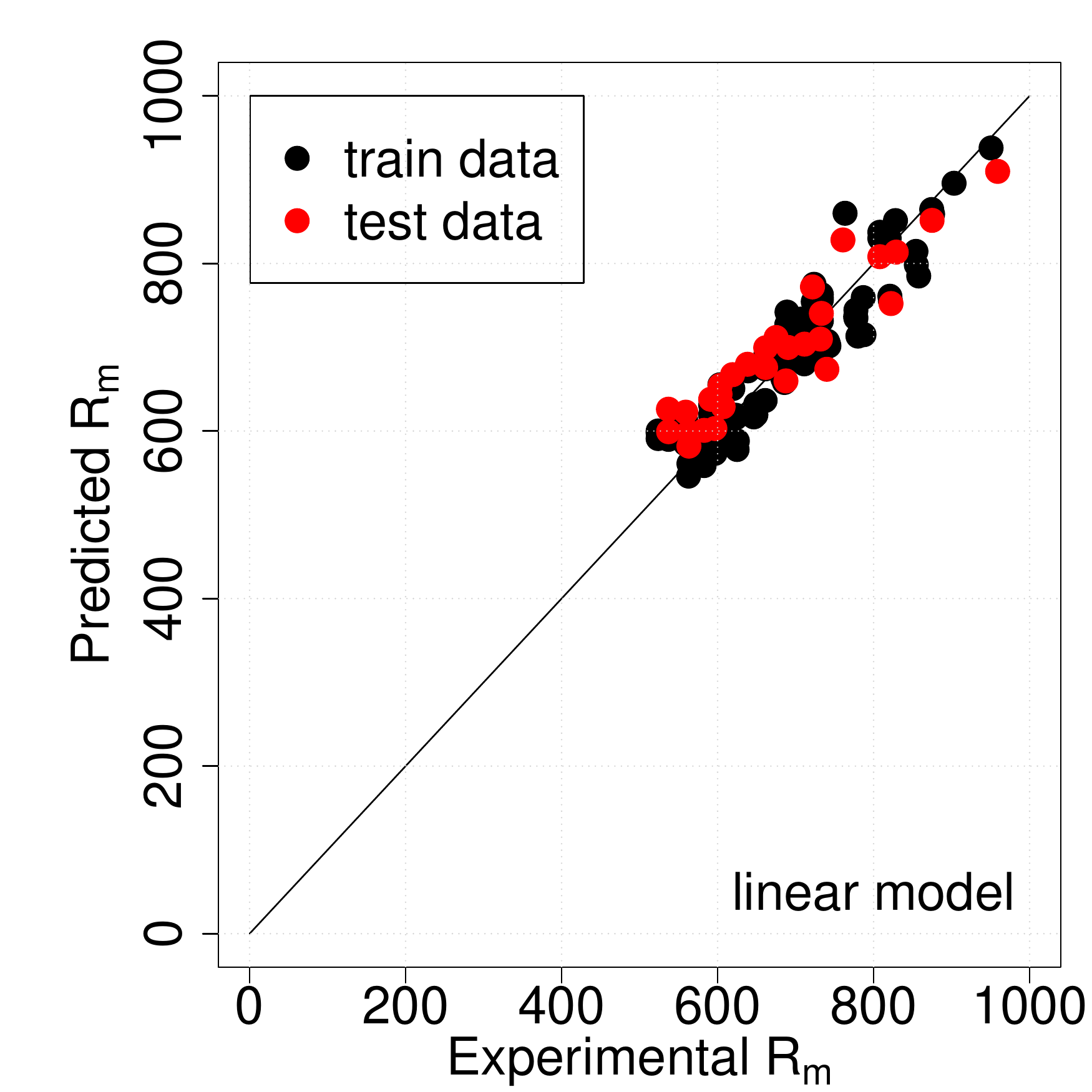}
          \put(-1,84){(b)}
      \end{overpic}
   \end{minipage}
   \hspace{.05\linewidth}
   \begin{minipage}[b]{.3\linewidth} 
   \begin{overpic}[scale=0.3,,tics=10]
          {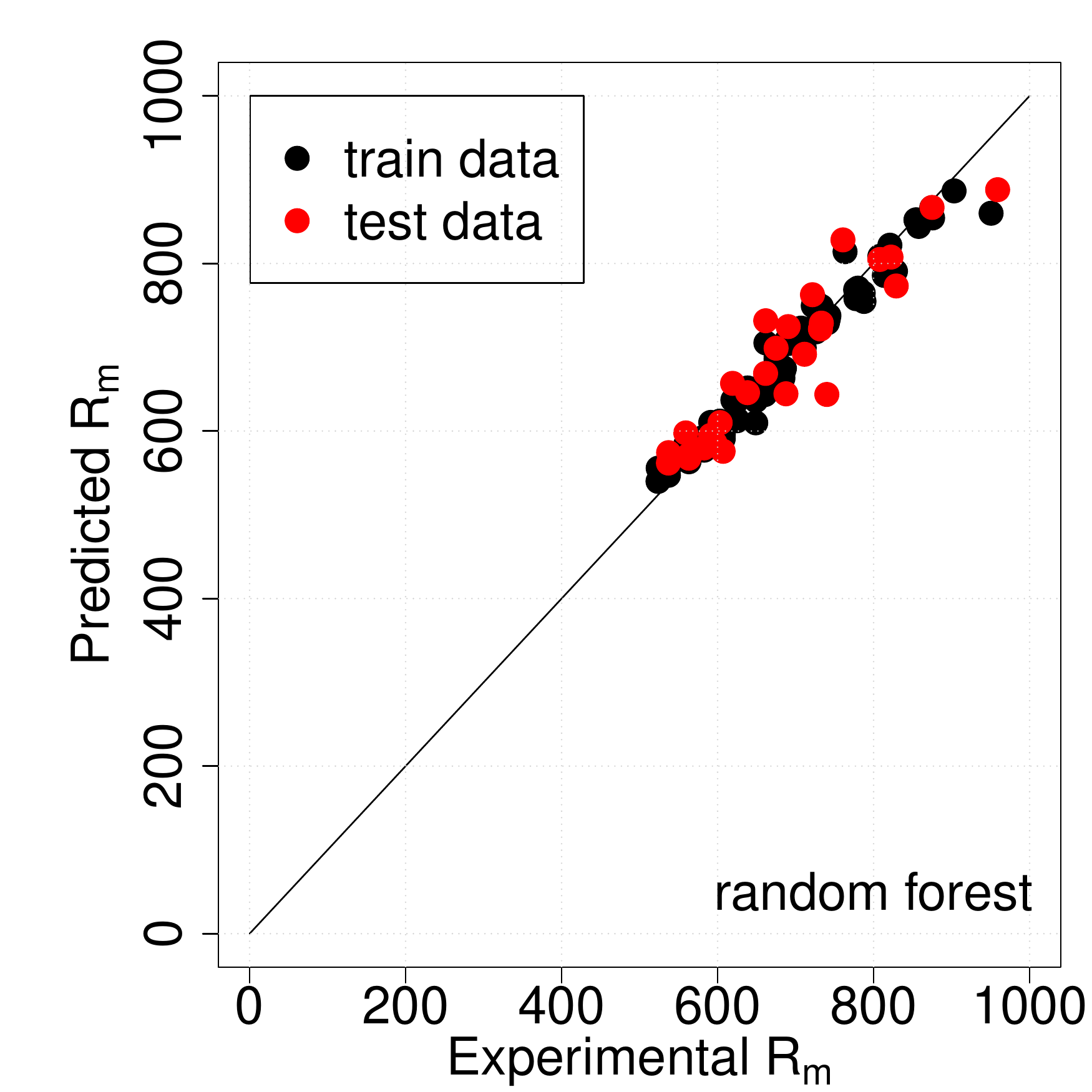}
          \put(-1,84){(c)}
      \end{overpic}
   \end{minipage}
   \caption{(a) Empirical $R_m$ vs. experimental $R_m$, (b) LM based $R_m$ vs.  experimental $R_m$, (c) RF based $R_m$ vs.  experimental $R_m$.}
   \label{model_compare}
\end{figure}

For the evaluation of the performance of the ML approaches 10-fold cross validation was conducted with the results shown in Table \ref{RMSE_comparison}.
\begin{table}
\centering
\begin{tabular}{ llll } 
\hline
  & empirical & LM & RF\\
 \hline
 RMSE & 44.1 & 41.2 & 40.4\\
 \hline
\end{tabular}
\caption{RMSE comparison.}
\label{RMSE_comparison}
\end{table}
Figures \ref{model_compare}(a-c) show the accuracy of the empirical approach, LM and RF, respectively.

\subsection{Application conclusions}
The application presented here aims at the determination of the tensile property $R_m$ from SPT data. Altstadt et. al \cite{altstadt2018estimation} proposed an empirical equation which correlates the punch force at onset of plastic instability of the SPT with $R_m$. The correlation factor $\beta$ for the estimation of $R_m$ needs to be determined before the empirical equation can be applied and $\beta$ is dependent on the sample geometry and the investigated material.
The application of ML models on SPT data works well for the prediction of $R_m$, as Table \ref{RMSE_comparison}shows. However, the amount of available data is very limited and the generalization power of the LM and RF for other materials, temperatures, geometries has to be critically evaluated.
RFs usually do not generalize well. Thus, the application of RF on SPT data might not be sensible for $R_m<500$ MPa or $R_m>1000$ MPa. More experimental data will have to be collected to verify the RF model further. However, the interpretation of the RF based on SPT data is straightforward and leads to relevant information about the importance of features for the performance of the model.

The LM approach is quite similar to the empirical approach of Altstadt et al. However, not only one specific force value of the SPT is used to correlate to $R_m$ but the combination of several principal components generated from the SPT curves are employed which might justify the slightly higher accuracy of the LM approach.
Thus, LM is a great tool to analyze the relationships among the variables but in combination with PCA the interpretability of the resulting model is not intuitive.

Overall, correlations between SPT curves and $R_m$ clearly exist and SPT data can be used for the prediction of tensile properties such as $R_m$ for five different steels (nine heats). Once the model is established, there is no further need for the conduction of time consuming tensile tests. In the future, the models might have to be re-established on a broader database. For the determination of $R_m$, FEM-simulations might not be necessary anymore. However, for more detailed information about the underlying constitutive laws via inverse analysis, FEM simulations will likely remain important an inevitable.

\section{Discussion and Conclusions}

This  survey  addresses  applications  of  machine  learning  strategies  in  materials  science for material characterization. 
There exists a wide range of promising applications for ML in materials science, e.g. material discovery, molecular dynamics and global structural prediction. The demand for new approaches dealing with limited data is huge.

It was shown that data-driven approaches play a significant role in materials research in order to find relationships between the structure of a material and its properties. These  relationships  are  often  not  linear.   It  is
difficult to find generic patterns among multiple length scales and timescales.  With experiments only, this cannot be achieved \cite{rajan2005materials}.  Therefore, data-mining techniques are indispensable for the recognition of correlations in the (experimental and simulated) data. As the amount of publicly available materials data grows, ML techniques in particular will be able to extract from these data sets
scientific principles and design rules that could not be determined through conventional analysis \cite{jain2016new}.

The majority of early ML applications to  materials science employed straightforward and simple-to-use algorithms, like linear kernel models and DTs. Now, these proofs-of-concept exist for a variety of application even though there is a lack of benchmarking datasets and standards \cite{schmidt2019recent}. 
To date, ML often cannot realize the expected accuracy when applied to some tasks due to insufficient material data. Therefore, a more accurate model that was trained on a small but accurate data set is only meaningful within the input data space but does not generalize well while a less accurate model on a wide input data space is better at generalization but less exact. 
Therefore, accelerating the construction of publicly accessible material databases is highly important for the future development of ML in materials science\cite{wei2019machine}. 
Another issue that holds back the development of precise ML models is the absence of failure data.  In this case, a cultural shift toward the publication of all valid data, may it be positive or negative, is required \cite{schmidt2019recent, wei2019machine}.

The majority of ML approaches in materials science is based on ANNs. However, conventional ANNs still suffer from several weaknesses such as the need for a large number of controlling parameters, the difficulty in obtaining stable solutions, the danger of overfitting and thus the lack of generalization capability \cite{fang2008hybrid}. However, ANNs have been enormously successful in understanding complex materials  behavior,  such  as  mechanical  behavior  (flow  stress,  hardness,  tensile strength, fracture strength, and fatigue behavior) of metal alloys subjected to certain heat treatment and/or deformation procedures, as well as in the prediction of micro-structures  and  phases  resulting  from  heat  treatment  and/or  deformation  processes. The most practical way to capture the complex dependence of a desired macroscopic property  on  the  various  process  parameters  is  through  such  learning  methods \cite{mueller2016machine}. ANNs have the potential to minimize the need for expensive  experimental  investigation  and/or  inspection  of  structural materials used  in  various  applications,  hence  resulting  in  large  economic benefits for organisations \cite{sterjovski2005artificial}.

In addition to ANN hybrid ML models or ensemble methods work well \cite{fang2008hybrid}. For this, multiple independent models are built and the final regression or classification result is usually obtained as an average over the ensemble. In this way, additional noise is introduced into the fitting process and overfitting is avoided  \cite{schmidt2019recent}. However, there does not exist  an overall solution that can be considered the best. The most appropriate model has always to be found specifically for the application and data situation.

\vspace{0.3cm}
Furthermore, this paper focuses on ML based material property prediction from SPT data. Such simple material tests have gained popularity over the last couple of years because even though they are cheap and simple to perform, they make accurate material characterization possible, especially for failure analysis and remaining life assessment of in-service components or structural parts. Nevertheless, a few disadvantages have to be taken into consideration. The small sample size might not represent the bulk material; the sample size effect influences the material properties; and the results of the SPT are sensitive to test parameters. 
However, also for SPT data, ML based model are popular for material parameter prediction. Most commonly found are ANNs, especially in combination with FEM for data generation. No application of traditional ML models to SPT data was found in the literature.
The paper concludes with an application example which uses FDCs of structural materials as the basis for predicting the UTS. Simple ML approaches presented here, such as linear regression models or RFs provide good results for predicting the UTS based on SPT data, even for a very small database.

As a consequence, it is possible to confirm the benefit
of simple ML techniques in predicting mechanical properties such as the UTS based on as simple material tests as the SPT and the authors are sure that ML will positively shape materials science for the years to come.

\section*{Acknowledgments}
This work is part of the Fraunhofer Lighthouse Project ML4P (Machine Learning for Production). The  work  profited  from  BiGmax, the Max Planck Society's Research Network on Big-Data-Driven Materials-Science.
The data from Section 3 was kindly provided by Dr. Altstadt of the Helmholtz-Zentrum Dresden-Rossendorf, Germany.


\subsection*{Financial disclosure}

None reported.

\subsection*{Conflict of interest}

The authors declare no potential conflict of interests.


%



\bibliography{GAMM_paper_new}

\providecommand{\url}[1]{\texttt{#1}}
\providecommand{\urlprefix}{}
\providecommand{\foreignlanguage}[2]{#2}
\providecommand{\Capitalize}[1]{\uppercase{#1}}
\providecommand{\capitalize}[1]{\expandafter\Capitalize#1}
\providecommand{\bibliographycite}[1]{\cite{#1}}
\providecommand{\bbland}{and}
\providecommand{\bblchap}{chap.}
\providecommand{\bblchapter}{chapter}
\providecommand{\bbletal}{et~al.}
\providecommand{\bbleditors}{editors}
\providecommand{\bbleds}{eds: }
\providecommand{\bbleditor}{editor}
\providecommand{\bbled}{ed.}
\providecommand{\bbledition}{edition}
\providecommand{\bbledn}{ed.}
\providecommand{\bbleidp}{page}
\providecommand{\bbleidpp}{pages}
\providecommand{\bblerratum}{erratum}
\providecommand{\bblin}{in}
\providecommand{\bblmthesis}{Master's thesis}
\providecommand{\bblno}{no.}
\providecommand{\bblnumber}{number}
\providecommand{\bblof}{of}
\providecommand{\bblpage}{page}
\providecommand{\bblpages}{pages}
\providecommand{\bblp}{p}
\providecommand{\bblphdthesis}{Ph.D. thesis}
\providecommand{\bblpp}{pp}
\providecommand{\bbltechrep}{}
\providecommand{\bbltechreport}{Technical Report}
\providecommand{\bblvolume}{volume}
\providecommand{\bblvol}{Vol.}
\providecommand{\bbljan}{January}
\providecommand{\bblfeb}{February}
\providecommand{\bblmar}{March}
\providecommand{\bblapr}{April}
\providecommand{\bblmay}{May}
\providecommand{\bbljun}{June}
\providecommand{\bbljul}{July}
\providecommand{\bblaug}{August}
\providecommand{\bblsep}{September}
\providecommand{\bbloct}{October}
\providecommand{\bblnov}{November}
\providecommand{\bbldec}{December}
\providecommand{\bblfirst}{First}
\providecommand{\bblfirsto}{1st}
\providecommand{\bblsecond}{Second}
\providecommand{\bblsecondo}{2nd}
\providecommand{\bblthird}{Third}
\providecommand{\bblthirdo}{3rd}
\providecommand{\bblfourth}{Fourth}
\providecommand{\bblfourtho}{4th}
\providecommand{\bblfifth}{Fifth}
\providecommand{\bblfiftho}{5th}
\providecommand{\bblst}{st}
\providecommand{\bblnd}{nd}
\providecommand{\bblrd}{rd}
\providecommand{\bblth}{th}
\begin{thebibliography}{1}

\bibitem{Rothermel1997}
G.~Rothermel, {\it ACM {T}ransactions on {S}oftware {E}ngineering and
  {M}ethodology} \textbf{1997}, {\it 6} (2), 173--210.

\bibitem{Rothermel1998}
G.~Rothermel, M.~J. Harrold, C.~W. Hirt, A.~A. Amsden, J.~L. Cook, {\it {ACM}
  {T}ransactions on {S}oftware {E}ngineering and {M}ethodology} \textbf{1998},
  {\it 6} (2), 173--210.

\bibitem{Elbaum2002}
S.~Elbaum, A.~G. Malishevsky, G.~Rothermel, {\it IEEE {T}ransactions on
  {S}oftware {E}ngineering} \textbf{February 2002}, {\it 28}, 159--182.

\bibitem{Allen2011}
T.~T. Allen, {\it Introduction to {D}iscrete {E}vent {S}imulation and
  {A}gent-based {M}odeling: {V}oting {S}ystems{,} {H}ealth {C}are{,}
  {M}ilitary{,} and {M}anufacturing}, Springer, New York, \textbf{2011}.

\bibitem{Yoo2007}
{A. L. Gould}, \bblin{} {\it Proceedings of Biopharmaceutical Section},
  American Statistical Association, Washington, D.C., \textbf{1983}, \bblpp{}.
  31--34.

\bibitem{Schulz2012}
A.~Schulz, G.~Doblhammer, \bblin{} {\it Versorgungs-Report},
  (\capitalize\bbleds{}C.~Gnster, J.~Klose, N.~Schmacke ), IEEE {P}ress,
  Piscataway, NJ, USA, \textbf{2012}, \bblpp{}. 161--175.

\bibitem{Blanchard2015}
G.~Blanchard, R.~Loubere, {\it High-Order {C}onservative {R}emapping with a
  posteriori {MOOD} stabilization on polygonal meshes}, \textbf{2015},
  Available from: \url{http://www.emn.fr/z-info/choco-solver/} [last accessed
  {M}ay 2011].

\bibitem{Ballen2011}
T.~T. Ballen, {\it Introduction to {D}iscrete {E}vent {S}imulation and
  {A}gent-based {M}odeling: {V}oting {S}ystems{,} {H}ealth {C}are{,}
  {M}ilitary{,} and {M}anufacturing} {\it \bblsecondo{} \bbledn{}}, Springer,
  New York, \textbf{2011}.

\bibitem{Boggs2002}
R.~Boggs, J.~Bozman, R.~Perry, Reducing downtime and business loss: {A}dressing
  business risk with effective technology, {\it \bbltechrep{} Technical report
  91-18}, International {D}ata {C}orporation ({IDC}), Sernageomin,
  \textbf{2002}.

\end{thebibliography}


\providecommand{\url}[1]{\texttt{#1}}
\providecommand{\urlprefix}{}
\providecommand{\foreignlanguage}[2]{#2}
\providecommand{\Capitalize}[1]{\uppercase{#1}}
\providecommand{\capitalize}[1]{\expandafter\Capitalize#1}
\providecommand{\bibliographycite}[1]{\cite{#1}}
\providecommand{\bbland}{and}
\providecommand{\bblchap}{chap.}
\providecommand{\bblchapter}{chapter}
\providecommand{\bbletal}{et~al.}
\providecommand{\bbleditors}{editors}
\providecommand{\bbleds}{eds: }
\providecommand{\bbleditor}{editor}
\providecommand{\bbled}{ed.}
\providecommand{\bbledition}{edition}
\providecommand{\bbledn}{ed.}
\providecommand{\bbleidp}{page}
\providecommand{\bbleidpp}{pages}
\providecommand{\bblerratum}{erratum}
\providecommand{\bblin}{in}
\providecommand{\bblmthesis}{Master's thesis}
\providecommand{\bblno}{no.}
\providecommand{\bblnumber}{number}
\providecommand{\bblof}{of}
\providecommand{\bblpage}{page}
\providecommand{\bblpages}{pages}
\providecommand{\bblp}{p}
\providecommand{\bblphdthesis}{Ph.D. thesis}
\providecommand{\bblpp}{pp}
\providecommand{\bbltechrep}{}
\providecommand{\bbltechreport}{Technical Report}
\providecommand{\bblvolume}{volume}
\providecommand{\bblvol}{Vol.}
\providecommand{\bbljan}{January}
\providecommand{\bblfeb}{February}
\providecommand{\bblmar}{March}
\providecommand{\bblapr}{April}
\providecommand{\bblmay}{May}
\providecommand{\bbljun}{June}
\providecommand{\bbljul}{July}
\providecommand{\bblaug}{August}
\providecommand{\bblsep}{September}
\providecommand{\bbloct}{October}
\providecommand{\bblnov}{November}
\providecommand{\bbldec}{December}
\providecommand{\bblfirst}{First}
\providecommand{\bblfirsto}{1st}
\providecommand{\bblsecond}{Second}
\providecommand{\bblsecondo}{2nd}
\providecommand{\bblthird}{Third}
\providecommand{\bblthirdo}{3rd}
\providecommand{\bblfourth}{Fourth}
\providecommand{\bblfourtho}{4th}
\providecommand{\bblfifth}{Fifth}
\providecommand{\bblfiftho}{5th}
\providecommand{\bblst}{st}
\providecommand{\bblnd}{nd}
\providecommand{\bblrd}{rd}
\providecommand{\bblth}{th}
\begin{thebibliography}{100}

\bibitem{agrawal2016perspective}
A.~Agrawal, A.~Choudhary, {\it Apl Materials} \textbf{2016}, {\it 4} (5),
  053208.

\bibitem{mueller2016machine}
T.~Mueller, A.~G. Kusne, R.~Ramprasad, {\it Rev Comput Chem} \textbf{2016},
  {\it 29}, 186--273.

\bibitem{wagner2016theory}
N.~Wagner, J.~M. Rondinelli, {\it Front Mater} \textbf{2016}, {\it 3}, 28.

\bibitem{dimiduk2018perspectives}
D.~M. Dimiduk, E.~A. Holm, S.~R. Niezgoda, {\it Integr Mater Manuf Innov}
  \textbf{2018}, {\it 7} (3), 157--172.

\bibitem{draxl2018nomad}
C.~Draxl, M.~Scheffler, {\it MRS Bull} \textbf{2018}, {\it 43} (9), 676--682.

\bibitem{verleysen2005curse}
M.~Verleysen, D.~Fran{\c{c}}ois, \bblin{} {\it International Work-Conference on
  Artificial Neural Networks}, Springer, \textbf{2005}, \bblpp{}. 758--770.

\bibitem{shen2019physical}
C.~Shen, C.~Wang, X.~Wei, Y.~Li, S.~van~der Zwaag, W.~Xu, {\it Acta Mater}
  \textbf{2019}, {\it 179}, 201--214.

\bibitem{arunkumar2016estimation}
S.~Arunkumar, Raghu~V. Prakash, {\it Trans Indian Inst Met} \textbf{2015}, {\it
  69} (6), 1245--1256.

\bibitem{butler2018machine}
K.~T. Butler, D.~W. Davies, H.~Cartwright, O.~Isayev, A.~Walsh, {\it Nature}
  \textbf{2018}, {\it 559} (7715), 547--555.

\bibitem{jordan2015machine}
M.I. Jordan, T.M. Mitchell, {\it Science} \textbf{2015}, {\it 349} (6245),
  255--260.

\bibitem{fan2014challenges}
J.~Fan, F.~Han, H.~Liu, {\it Natl Sci Rev} \textbf{2014}, {\it 1} (2),
  293--314.

\bibitem{national1974materials}
National~Academy of~Sciences (US). Committee on the Survey~of
  Materials~Science, National~Academy of~Sciences (US). Committee~on Science,
  Public Policy, {\it Materials and Man's Needs: {Materials} Science and
  Engineering; Supplementary Report}, {\it \bblvol{}~1}, National Academies,
  \textbf{1974}.

\bibitem{olson1997computational}
G.B. Olson, {\it Science} \textbf{1997}, {\it 277} (5330), 1237--1242.

\bibitem{correa2018accelerating}
J.~Correa-Baena, K.~Hippalgaonkar, J.~van Duren, S.~Jaffer, V.~R.
  Chandrasekhar, V.~Stevanovic, C.s Wadia, S.~Guha, T.~Buonassisi, {\it Joule}
  \textbf{2018}, {\it 2} (8), 1410--1420.

\bibitem{bock2019review}
F.~E. Bock, R.~C. Aydin, C.~J. Cyron, N.~Huber, S.~R. Kalidindi, B.~Klusemann,
  {\it Front Mater} \textbf{2019}, {\it 6}, 110.

\bibitem{wei2019machine}
J.~Wei, X.~Chu, X.~Y. Sun, K.~Xu, H.~X. Deng, J.~Chen, Z.~Wei, M.~Lei, {\it
  InfoMat} \textbf{2019}, {\it 1} (3), 338--358.

\bibitem{seko2017representation}
A.~Seko, H.~Hayashi, K.~Nakayama, A.~Takahashi, I.~Tanaka, {\it Phys Rev B}
  \textbf{2017}, {\it 95} (14), 144110.

\bibitem{schutt2014represent}
K.T. Schütt, H.~Glawe, F.~Brockherde, A.~Sanna, K.R. Müller, E.K.U. Gross,
  {\it Phys Rev B} \textbf{2014}, {\it 89} (20), 205118.

\bibitem{isayev2015materials}
O.~Isayev, D.~Fourches, E.~N. Muratov, C.~Oses, K.~Rasch, A.~Tropsha,
  S.~Curtarolo, {\it Chem Mater} \textbf{2015}, {\it 27} (3), 735--743.

\bibitem{medasani2016predicting}
B.~Medasani, A.~Gamst, H.~Ding, W.~Chen, K.~A. Persson, M.~Asta, A.~Canning,
  M.~Haranczyk, {\it npj Comput Mater} \textbf{2016}, {\it 2} (1), 1--10.

\bibitem{de2016statistical}
M.~de~Jong, W.~Chen, R.~Notestine, K.~Persson, G.~Ceder, A.~Jain, M.~Asta,
  A.~Gamst, {\it Sci Rep} \textbf{2016}, {\it 6} (1), 34256.

\bibitem{legrain2017chemical}
F.~Legrain, J.~Carrete, A.~van Roekeghem, S.~Curtarolo, N.~Mingo, {\it Chem
  Mater} \textbf{2017}, {\it 29} (15), 6220--6227.

\bibitem{li2015molecular}
Z.~Li, J.~R. Kermode, A.~De~Vita, {\it Phys Rev Lett} \textbf{2015}, {\it 114}
  (9), 096405.

\bibitem{li2017high}
Z.~Li, S.~Wang, W.~S. Chin, L.~E. Achenie, H.~Xin, {\it J Mater Chem A}
  \textbf{2017}, {\it 5} (46), 24131--24138.

\bibitem{ma2015machine}
X.~Ma, Z.~Li, L.~E.K. Achenie, H.~Xin, {\it J Phys Chem Lett} \textbf{2015},
  {\it 6} (18), 3528--3533.

\bibitem{mannodi2017mining}
A.~Mannodi-Kanakkithodi, T.~D. Huan, R.~Ramprasad, {\it Chem Mater}
  \textbf{2017}, {\it 29} (21), 9001--9010.

\bibitem{ulissi2017address}
Z.~W. Ulissi, A.~J. Medford, T.~Bligaard, J.~K. Nørskov, {\it Nat Commun}
  \textbf{2017}, {\it 8} (1), 1--7.

\bibitem{raccuglia2016machine}
P.~Raccuglia, K.~C. Elbert, P.~D.F. Adler, C.~Falk, M.~B. Wenny, A.~Mollo,
  M.~Zeller, S.~A. Friedler, J.~Schrier, A.~J. Norquist, {\it Nature}
  \textbf{2016}, {\it 533} (7601), 73--76.

\bibitem{kim2017virtual}
E.~Kim, K.~Huang, S.~Jegelka, E.~Olivetti, {\it npj Comput Mater}
  \textbf{2017}, {\it 3} (1), 1--9.

\bibitem{graser2018machine}
J.~Graser, S.~K. Kauwe, T.~D. Sparks, {\it Chem Mater} \textbf{2018}, {\it 30}
  (11), 3601--3612.

\bibitem{schmidt2019recent}
J.~Schmidt, M.~R.G. Marques, S.~Botti, M.~A.L. Marques, {\it npj Comput Mater}
  \textbf{2019}, {\it 5} (1), 1--36.

\bibitem{wold1987principal}
S.~Wold, K.~Esbensen, P.~Geladi, {\it Chemom Intell Lab Syst} \textbf{1987},
  {\it 2} (1-3), 37--52.

\bibitem{jolliffe2016principal}
I.~T. Jolliffe, J.~Cadima, {\it Phil Trans R Soc A} \textbf{2016}, {\it 374}
  (2065), 20150202.

\bibitem{vellido2012making}
A.~Vellido, J.~D. Mart{\'\i}n-Guerrero, P.~J.G. Lisboa, \bblin{} {\it ESANN},
  Citeseer, \textbf{2012}, \bblpp{}. 163--172.

\bibitem{erhan2009visualizing}
D.~Erhan, Y.~Bengio, A.~Courville, P.~Vincent, {\it University of Montreal}
  \textbf{2009}, {\it 1341} (3), 1.

\bibitem{simonyan2013deep}
K.~Simonyan, A.~Vedaldi, A.~Zisserman, {\it arXiv preprint arXiv:1312.6034}
  \textbf{2013}.

\bibitem{fong2017interpretable}
R.C. Fong, A.~Vedaldi, \bblin{} {\it Proceedings of the IEEE International
  Conference on Computer Vision}, \textbf{2017}, \bblpp{}. 3429--3437.

\bibitem{zeiler2014visualizing}
M.D. Zeiler, R.~Fergus, \bblin{} {\it European Conference on Computer Vision},
  Springer, \textbf{2014}, \bblpp{}. 818--833.

\bibitem{sundararajan2017axiomatic}
M.~Sundararajan, A.~Taly, Q.~Yan, {\it arXiv preprint arXiv:1703.01365}
  \textbf{2017}.

\bibitem{zhang2018strategy}
Y.~Zhang, C.~Ling, {\it npj Comput Mater} \textbf{2018}, {\it 4} (1), 1--8.

\bibitem{dey2014informatics}
P.~Dey, J.~Bible, S.~Datta, S.~Broderick, J.~Jasinski, M.~Sunkara, M.~Menon,
  K.~Rajan, {\it Comp Mater Sci} \textbf{2014}, {\it 83}, 185--195.

\bibitem{pilania2016machine}
G.~Pilania, A.~Mannodi-Kanakkithodi, B.P. Uberuaga, R.~Ramprasad, J.E.
  Gubernatis, T.~Lookman, {\it Sci Rep} \textbf{2016}, {\it 6} (1), 19375.

\bibitem{von2020informed}
L.~von Rueden, S.~Mayer, K.~Beckh, B.~Georgiev, S.~Giesselbach, R.~Heese,
  B.~Kirsch, J.~Pfrommer, A.~Pick, R.~Ramamurthy, \bbletal{}, {\it arXiv
  preprint arXiv:1903.12394} \textbf{2020}.

\bibitem{raissi2017physics}
M.~Raissi, P.~Perdikaris, G.~Karniadakis, {\it arXiv preprint arXiv:1711.10561}
  \textbf{2017}.

\bibitem{zhu2019physics}
Y.~Zhu, N.~Zabaras, P.S. Koutsourelakis, P.~Perdikaris, {\it J Comp Phy}
  \textbf{2019}, {\it 394}, 56--81.

\bibitem{zhang2020physics}
E.~Zhang, M.~Yin, G.E. Karniadakis, {\it arXiv preprint arXiv:2009.04525}
  \textbf{2020}.

\bibitem{struyf2007clustering}
J.~Struyf, S.~D{\v{z}}eroski, \bblin{} {\it European Conference on Machine
  Learning}, Springer, \textbf{2007}, \bblpp{}. 359--370.

\bibitem{ermon2015pattern}
S.~Ermon, R.L. Bras, S.K. Suram, J.M. Gregoire, C.~Gomes, B.~Selman, R.B.
  Van~Dover, {\it arXiv preprint arXiv:1411.7441} \textbf{2014}.

\bibitem{macinnes2009visual}
J.~MacInnes, S.~Santosa, W.~Wright, {\it IEEE Comput Grap Appl} \textbf{2010},
  {\it 30} (1), 8--14.

\bibitem{chang2007guiding}
M.~W. Chang, L.~Ratinov, D.~Roth, \bblin{} {\it Proceedings of the 45th Annual
  Meeting of the Association of Computational Linguistics}, \textbf{2007},
  \bblpp{}. 280--287.

\bibitem{dette2006simple}
H.~Dette, N.~Neumeyer, K.F. Pilz, {\it Bernoulli} \textbf{2006}, {\it 12} (3),
  469--490.

\bibitem{mgi}
The Materials~Genome Initiative, \url {https://www.mgi.gov/}, Accessed:
  2020-07-21.

\bibitem{mp}
Materials Project, \url {https://materialsproject.org/}, Accessed: 2020-07-21.

\bibitem{jain2013commentary}
A.~Jain, S.~P. Ong, G.~Hautier, W.~Chen, W.~D. Richards, S.~Dacek, S.~Cholia,
  D.~Gunter, D.~Skinner, G.~Ceder, K.~A. Persson, {\it APL Mater}
  \textbf{2013}, {\it 1} (1), 011002.

\bibitem{icsd}
S.~Rühl, {\it The Inorganic Crystal Structure Database ( {ICSD} ): {A} Tool
  for Materials Sciences}, \url {https://icsd.products.fiz-karlsruhe.de/},
  \textbf{2019}, Accessed: 2020-07-21.

\bibitem{international1987crystallographic}
G.~Gergerhoff F.~H.~Allen, R.~Sievers, {\it Crystallographic databases:
  {Information} content, software systems, scientific applications},
  International Union of Crystallography. Data Commission, \textbf{1987}.

\bibitem{white2012materials}
A.~White, {\it MRS Bull} \textbf{2012}, {\it 37} (8), 715--716.

\bibitem{nr}
Nomad Repository, \url {https://nomad-coe.eu/}, Accessed: 2020-07-21.

\bibitem{topo_db}
Topological~Materials Database, \url
  {https://www.topologicalquantumchemistry.com/}, Accessed: 2020-09-21.

\bibitem{supercon}
Materials Information~Station National Institute~of Materials~Science, {\it
  {SuperCon}}, \url {https://supercon.nims.go.jp/en/}, Accessed: 2020-07-21.

\bibitem{nims}
National~Institute of~Materials~Science, \url {https://www.nims.go.jp/eng/},
  Accessed: 2020-07-21.

\bibitem{nist}
NIST Materials~Data Repository, \url {https://materialsdata.nist.gov/},
  \textbf{2020}, Accessed: 2020-07-21.
  \urlprefix\url{https://doi.org/10.1002/9783527809080.cataz11525}.

\bibitem{saal2013materials}
J.~E. Saal, S.~Kirklin, M.~Aykol, B.~Meredig, C.~Wolverton, {\it JOM}
  \textbf{2013}, {\it 65} (11), 1501--1509.

\bibitem{kirklin2015open}
S.~Kirklin, J.~E. Saal, B.~Meredig, A.~Thompson, J.~W. Doak, M.~Aykol,
  S.~Rühl, C.~Wolverton, {\it npj Comput Mater} \textbf{2015}, {\it 1} (1),
  1--15.

\bibitem{groom2016cambridge}
C.~R. Groom, I.~J. Bruno, M.~P. Lightfoot, S.~C. Ward, {\it Acta Crystallogr
  Sect B} \textbf{2016}, {\it 72} (2), 171--179.

\bibitem{puchala2016materials}
B.~Puchala, G.~Tarcea, E.~A. Marquis, M.~Hedstrom, H.V. Jagadish, J.~E.
  Allison, {\it JOM} \textbf{2016}, {\it 68} (8), 2035--2044.

\bibitem{Zakutayev2018open}
A.~Zakutayev, N.~Wunder, M.~Schwarting, J.~D. Perkins, R.~White, K.~Munch,
  W.~Tumas, C.~Phillips, {\it Sci Data} \textbf{2018}, {\it 5} (1), 180053.

\bibitem{ward2018strategies}
L.~Ward, M.~Aykol, B.~Blaiszik, I.~Foster, B.~Meredig, J.~Saal, S.~Suram, {\it
  MRS Bull} \textbf{2018}, {\it 43} (9), 683--689.

\bibitem{mlinaric2017dealing}
A.~Mlinarić, M.~Horvat, V.~Šupak Smolčić, {\it Biochem Medica}
  \textbf{2017}, {\it 27} (3), 447--452.

\bibitem{liu2017materials}
Y.~Liu, T.~Zhao, W.~Ju, S.~Shi, {\it J Materiomics} \textbf{2017}, {\it 3} (3),
  159--177.

\bibitem{song2004preliminary}
Q.~Song, {\it Chinese Sci Bull} \textbf{2004}, {\it 49} (2), 210.

\bibitem{wei2006materials}
Q.~Wei, X.~Peng, X.~Liu, W.~Xie, {\it Chinese Sci Bull} \textbf{2006}, {\it 51}
  (4), 498--504.

\bibitem{blaiszik2016materials}
B.~Blaiszik, K.~Chard, J.~Pruyne, R.~Ananthakrishnan, S.~Tuecke, I.~Foster,
  {\it JOM} \textbf{2016}, {\it 68} (8), 2045--2052.

\bibitem{o2016materials}
J.~O’Mara, B.~Meredig, K.~Michel, {\it JOM} \textbf{2016}, {\it 68} (8),
  2031--2034.

\bibitem{gaultois2016perspective}
M.~W. Gaultois, A.~O. Oliynyk, A.~Mar, T.~D. Sparks, G.~J. Mulholland,
  B.~Meredig, {\it APL Mater} \textbf{2016}, {\it 4} (5), 053213.

\bibitem{rajan2005materials}
K.~Rajan, {\it Mater Today} \textbf{2005}, {\it 8} (10), 38--45.

\bibitem{lucas2014connecting}
A.~Lucas, {\it arXiv preprint arXiv:1402.2593} \textbf{2014}.

\bibitem{meyers2008mechanical}
M.~A. Meyers, K.~K. Chawla, {\it Mechanical Behavior of Materials}, Cambridge
  University Press, \textbf{2008}.

\bibitem{agarwala2000corrosion}
V.~S. Agarwala, P.~L. Reed, S.~Ahmad, \bbletal{}, \bblin{} {\it CORROSION
  2000}, NACE International, \textbf{2000}.

\bibitem{jimenez2014automatic}
M.J. Jiménez–Come, I.J. Turias, F.J. Trujillo, {\it Mater Design}
  \textbf{2014}, {\it 56}, 642--648.

\bibitem{atha2018evaluation}
D.~J. Atha, M.~R. Jahanshahi, {\it Struct Health Monit} \textbf{2017}, {\it 17}
  (5), 1110--1128.

\bibitem{bastian2019visual}
B.~T. Bastian, Jaspreeth N., S.~K. Ranjith, C.V. Jiji, {\it NDT\&E Int}
  \textbf{2019}, {\it 107}, 102134.

\bibitem{fang2008hybrid}
S.F. Fang, M.P. Wang, W.H. Qi, F.~Zheng, {\it Comp Mater Sci} \textbf{2008},
  {\it 44} (2), 647--655.

\bibitem{hoang2019image}
N.~D. Hoang, V.~D. Tran, {\it Comput Intel Neurosc} \textbf{2019}, {\it 2019},
  1--13.

\bibitem{wen2009corrosion}
Y.F. Wen, C.Z. Cai, X.H. Liu, J.F. Pei, X.J. Zhu, T.T. Xiao, {\it Corros Sci}
  \textbf{2009}, {\it 51} (2), 349--355.

\bibitem{shiraiwa2017fatigue}
T.~Shiraiwa, F.~Briffod, Y.~Miyazawa, M.~Enoki, \bblin{} {\it Proceedings of
  the 4th World Congress on Integrated Computational Materials Engineering
  (ICME 2017)}, Springer, \textbf{2017}, \bblpp{}. 317--326.

\bibitem{agrawal2014exploration}
A.~Agrawal, P.~D. Deshpande, A.~Cecen, G.~P. Basavarsu, A.~N. Choudhary, S.~R.
  Kalidindi, {\it Integr Mater Manuf Innov} \textbf{2014}, {\it 3} (1),
  90--108.

\bibitem{zhang2015using}
L.~Zhang, J.~Lei, Q.~Zhou, Y.~Wang, {\it Adv in Nat Sci} \textbf{2015}, {\it 8}
  (1), 21--26.

\bibitem{abdalla2011modeling}
J.~A. Abdalla, R.~Hawileh, {\it J Franklin Inst} \textbf{2011}, {\it 348} (7),
  1393--1403.

\bibitem{chatzidakis2014creep}
S.~Chatzidakis, M.~Alamaniotis, L.~H. Tsoukalas, {\it IJMSTR} \textbf{2014},
  {\it 2} (2), 1--25.

\bibitem{frolova2011representation}
O.~Frolova, E.~Roos, K.~Maile, W.~M{\"u}ller, {\it Trans MLDM} \textbf{2011},
  {\it 4} (1), 1--16.

\bibitem{wang2019design}
C.~Wang, C.~Shen, X.~Huo, C.~Zhang, W.~Xu, {\it Nucl Eng Technol}
  \textbf{2020}, {\it 52} (5), 1008--1012.

\bibitem{sourmail2002neural}
T.~Sourmail, H.K.D.H. Bhadeshia, D.J.C. MacKay, {\it Mater Sci Technol}
  \textbf{2002}, {\it 18} (6), 655--663.

\bibitem{shin2019modern}
D.~Shin, Y.~Yamamoto, M.~P. Brady, S.~Lee, J.~Haynes, {\it SSRN Journal}
  \textbf{2018}, {\it 168}, 321--330.

\bibitem{gutscher2004determination}
G.~Gutscher, H.~C. Wu, G.~Ngaile, T.~Altan, {\it J Mater Process Technol}
  \textbf{2004}, {\it 146} (1), 1--7.

\bibitem{wright2016wire}
R.~N. Wright, {\it Wire Technology}, Elsevier, \textbf{2016}.

\bibitem{lin2008application}
Y.C. Lin, Jun Zhang, Jue Zhong, {\it Comp Mater Sci} \textbf{2008}, {\it 43}
  (4), 752--758.

\bibitem{gupta2013development}
A.~K. Gupta, H.~N. Krishnamurthy, Y.~Singh, K.~M. Prasad, S.~K. Singh, {\it
  Mater Design} \textbf{2013}, {\it 45}, 616--627.

\bibitem{desu2014support}
R.~K. Desu, S.~C. Guntuku, B.~Aditya, A.~K. Gupta, {\it Procedia Mater Sci}
  \textbf{2014}, {\it 6}, 368--375.

\bibitem{meng2015identification}
L.~Meng, P.~Breitkopf, B.~Raghavan, G.~Mauvoisin, O.~Bartier, X.~Hernot, {\it
  Comput Methods Appl Mech Eng} \textbf{2015}, {\it 297}, 239--257.

\bibitem{mahalle2019neural}
G.~Mahalle, O.~Salunke, N.~Kotkunde, A.~K. Gupta, S.~K. Singh, {\it J Mater Res
  Technol} \textbf{2019}, {\it 8} (2), 2130--2140.

\bibitem{rahman2019tensile}
R.~Rahman, S.~Zhafer Firdaus Syed~Putra \bblin{} {\it Mechanical and Physical
  Testing of Biocomposites, Fibre-Reinforced Composites and Hybrid Composites},
  Elsevier, \textbf{2019}, \bblpp{}. 81--102.

\bibitem{wang2020tensile}
C.~Wang, C.~Shen, Q.~Cui, C.~Zhang, W.~Xu, {\it J Nucl Mater} \textbf{2020},
  {\it 529}, 151823.

\bibitem{santos2009machine}
I.~Santos, J.~Nieves, Y.~K. Penya, P.~G. Bringas, \bblin{} {\it 2009
  ICCAS-SICE}, IEEE, \textbf{2009}, \bblpp{}. 4536--4541.

\bibitem{sterjovski2005artificial}
Z.~Sterjovski, D.~Nolan, K.R. Carpenter, D.P. Dunne, J.~Norrish, {\it J Mater
  Process Technol} \textbf{2005}, {\it 170} (3), 536--544.

\bibitem{sankar2009applying}
I.Z. Sankar, K.M. Rao, B.V.R. Murhty, {\it IE (I) Journal-MM} \textbf{2009},
  {\it 90}, 3--6.

\bibitem{datta2007designing}
S.~Datta, F.~Pettersson, S.~Ganguly, H.~Saxén, N.~Chakraborti, {\it ISIJ Int}
  \textbf{2007}, {\it 47} (8), 1195--1203.

\bibitem{pattanayak2015computational}
S.~Pattanayak, S.~Dey, S.~Chatterjee, S.~G. Chowdhury, S.~Datta, {\it Comp
  Mater Sci} \textbf{2015}, {\it 104}, 60--68.

\bibitem{metzbower2001neural}
E.A. Metzbower, J.J. deLoach, S.H. Lalam, H.K.D.H. Bhadeshia, {\it Sci Technol
  Weld Joining} \textbf{2001}, {\it 6} (2), 116--124.

\bibitem{poudel2013selective}
R.~C. Poudel, T.~Sakaguchi, Y.~Shimizu, {\it J Chem Eng Japan} \textbf{2013},
  {\it 46} (4), 294--301.

\bibitem{shigemori2007optimum}
H.~Shigemori, S.~Kawamura, \bblin{} {\it SICE Annual Conference 2007}, IEEE,
  IEEE, \textbf{2007}, \bblpp{}. 810--815.
  \urlprefix\url{https://doi.org/10.1109/sice.2007.4421095}.

\bibitem{shigemori2011optimum}
H.~Shigemori, M.~Kano, S.~Hasebe, {\it J Process Control} \textbf{2011}, {\it
  21} (2), 293--301.

\bibitem{swaddiwudhipong2005material}
S.~Swaddiwudhipong, K.~K. Tho, Z.~S. Liu, J.~Hua, N.~S.~B. Ooi, {\it Modelling
  Simul Mater Sci Eng} \textbf{2005}, {\it 13} (6), 993--1004.

\bibitem{fragassa2019predicting}
C.~Fragassa, M.~Babic, C.~P. Bergmann, G.~Minak, {\it Metals} \textbf{2019},
  {\it 9} (5), 557.

\bibitem{christopher2018neural}
L.~Christopher, T.~Sasikumar, C.~Santulli, C.~Fragassa, {\it FME Transaction}
  \textbf{2018}, {\it 46} (3), 253--258.

\bibitem{sasikumar2008artificial}
T.~Sasikumar, S.~Rajendraboopathy, K.M. Usha, E.S. Vasudev, {\it J Nondestruct
  Eval} \textbf{2008}, {\it 27} (4), 127--133.

\bibitem{liu2012control}
Q.~Liu, X.~Zhang, B.~Wang, B.~Wang, {\it Sci Technol Cast Proc} \textbf{2012},
  {\it 26}.

\bibitem{kwon2019prediction}
S.~H. Kwon, D.~G. Hong, C.~H. Yim, {\it Ironmak Steelmak} \textbf{2019}, 1--12.

\bibitem{husain2017small}
A.~Husain, R.~Sharma, D.K. Sehgal, {\it Proc Eng} \textbf{2017}, {\it 173},
  710--717.

\bibitem{altstadt2018estimation}
E.~Altstadt, M.~Houska, I.~Simonovski, M.~Bruchhausen, S.~Holmström,
  R.~Lacalle, {\it Int J Mech Sci} \textbf{2018}, {\it 136}, 85--93.

\bibitem{wang2017determination}
H.~Wang, T.~Xu, B.~Shou, {\it Materials} \textbf{2016}, {\it 10} (1), 23.

\bibitem{manahan1981development}
M.P. Manahan, A.S. Argon, O.K. Harling, {\it J Nucl Mater} \textbf{1981}, {\it
  104}, 1545--1550.

\bibitem{manahan1982development}
M.~P. Manahan, \bblphdthesis{}, Massachusetts Institute of Technology,
  \textbf{1982}.

\bibitem{astm}
Standard Test~Method for Small Punch Testing of Ultra-High Molecular Weight
  Polyethylene Used~in Surgical~Implants, {\it {ASTM F2183-02(2008)}}, \url
  {https://www.astm.org/Standards/F2183.htm}, Accessed: 2020-07-21.

\bibitem{gbt}
Small punch test methods of metallic materials for in-service~pressure
  equipment, {\it {GB/T 29459}}, \url
  {https://www.chinesestandard.net/PDF.aspx/GBT29459.2-2012}, Accessed:
  2020-07-21.

\bibitem{cwa}
CEN CWA-15627, {\it 15627 Worskshop Agreement: Small punch test method for
  metallic materials}, \textbf{2006}.

\bibitem{arunkumar2020overview}
S.~Arunkumar, {\it Met Mater Int} \textbf{2019}, {\it 26} (6), 719--738.

\bibitem{janvca2016small}
A.~Jan{\v{c}}a, J.~Siegl, P.~Hau{\v{s}}ild, {\it J Nucl Mater} \textbf{2016},
  {\it 481}, 201--213.

\bibitem{kim2014comparative}
S.~W. Kim, Y.~S. Lee, {\it Metall and Materi Trans B} \textbf{2013}, {\it 45}
  (2), 445--453.

\bibitem{klevtsov2008using}
I.~Klevtsov, A.~Dedov, A.~Molodtsov, \bblin{} {\it 6th International DAAAM
  Baltic Conference Industrial Engineering}, \textbf{2008}.

\bibitem{n2013mechanical}
M.~Yetna~N’Jock, D.~Chicot, X.~Decoopman, J.~Lesage, J.M. Ndjaka, A.~Pertuz,
  {\it Int J Mech Sci} \textbf{2013}, {\it 75}, 257--264.

\bibitem{tabor1951hardness}
D.~Tabor, {\it J Inst Metals} \textbf{1951}, {\it 79}, 1.

\bibitem{meng2017objective}
L.~Meng, B.~Raghavan, O.~Bartier, X.~Hernot, G.~Mauvoisin, P.~Breitkopf, {\it
  Mech Mater} \textbf{2017}, {\it 107}, 31--44.

\bibitem{milivcka2006small}
K.~Milička, F.~Dobeš, {\it Int J Press Vessels Pip} \textbf{2006}, {\it 83}
  (9), 625--634.

\bibitem{chakrabarty1970theory}
J.~Chakrabarty, {\it Int J Mech Sci} \textbf{1970}, {\it 12} (4), 315--325.

\bibitem{byun2001characterization}
T.S. Byun, E.H. Lee, J.D. Hunn, K.~Farrell, L.K. Mansur, {\it J Nucl Mater}
  \textbf{2001}, {\it 294} (3), 256--266.

\bibitem{campitelli2004assessment}
E.N. Campitelli, P.~Spätig, R.~Bonadé, W.~Hoffelner, M.~Victoria, {\it J Nucl
  Mater} \textbf{2004}, {\it 335} (3), 366--378.

\bibitem{abendroth2006identification}
M.~Abendroth, M.~Kuna, {\it Eng Fract Mech} \textbf{2006}, {\it 73} (6),
  710--725.

\bibitem{simonovski2017small}
I.~Simonovski, S.~Holmström, M.~Bruchhausen, {\it Int J Mech Sci}
  \textbf{2017}, {\it 120}, 204--213.

\bibitem{linse2008usage}
T.~Linse, M.~Kuna, J.~Schuhknecht, H.-W. Viehrig, {\it Eng Fract Mech}
  \textbf{2008}, {\it 75} (11), 3520--3533.

\bibitem{lu2020extraction}
L.~Lu, M.~Dao, P.~Kumar, U.~Ramamurty, G.E. Karniadakis, S.~Suresh, {\it
  Proceedings of the National Academy of Sciences} \textbf{2020}, {\it 117}
  (13), 7052--7062.

\bibitem{penuelas2011inverse}
I.~Pe{\~n}uelas, C.~Beteg{\'o}n, C.~Rodr{\'\i}guez, J.~Belzunce, {\it Numerical
  Simulations-Applications, Examples and Theory, INCTECH, Rijeka (Croatia)}
  \textbf{2011}, 311--330.

\bibitem{kameda1986kinetic}
J.~Kameda, {\it Acta Metall} \textbf{1986}, {\it 34} (12), 2391--2398.

\bibitem{misawa1987small}
T.~Misawa, T.~Adachi, M.~Saito, Y.~Hamaguchi, {\it J Nucl Mater} \textbf{1987},
  {\it 150} (2), 194--202.

\bibitem{mcnaney1991application}
J.~McNaney, G.E. Lucas, G.R. Odette, {\it J Nucl Mater} \textbf{1991}, {\it
  179-181}, 429--433.

\bibitem{kameda1992small}
J.~Kameda, X.~Mao, {\it J Mater Sci} \textbf{1992}, {\it 27} (4), 983--989.

\bibitem{linse2014quantification}
T.~Linse, M.~Kuna, H.-W. Viehrig, {\it Mater Sci Eng A} \textbf{2014}, {\it
  614}, 136--147.

\bibitem{mao1987development}
X.~Mao, H.~Takahashi, {\it J Nucl Mater} \textbf{1987}, {\it 150} (1), 42--52.

\bibitem{hurst2012we}
R.~Hurst, K.~Matocha, {\it K. Matocha, R. Hurst, W. Sun, Determination of
  Mechanical Properties of Materials by Small Punch and other Miniature Testing
  Techniques. Ostrava, OCELOT sro} \textbf{2012}, 4--18.

\bibitem{oliver1992improved}
W.C. Oliver, G.M. Pharr, {\it J Mater Res} \textbf{1992}, {\it 7} (6),
  1564--1583.

\bibitem{field1993simple}
J.S. Field, M.V. Swain, {\it J Mater Res} \textbf{1993}, {\it 8} (2), 297--306.

\bibitem{taljat1997analysis}
B.~Taljat, T.~Zacharia, F.M. Haggag, {\it J Mater Res} \textbf{1997}, {\it 12}
  (4), 965--974.

\bibitem{huber1997determination}
N.~Huber, D.~Munz, Ch. Tsakmakis, {\it J Mater Res} \textbf{1997}, {\it 12}
  (9), 2459--2469.

\bibitem{konopik2013determination}
P.~Konop{\'\i}k, J.~Dzugan, R.~Prochazka, {\it Metal Brno Czech Republic}
  \textbf{2013}, {\it 2013} (5), 15--17.

\bibitem{wang2008small}
Z.~X. Wang, H.~J. Shi, J.~Lu, P.~Shi, X.~F. Ma, {\it Nucl Eng Des}
  \textbf{2008}, {\it 238} (12), 3186--3193.

\bibitem{bulloch1998toughness}
J.H Bulloch, {\it Int J Press Vessels Pip} \textbf{1998}, {\it 75} (11),
  791--804.

\bibitem{mao1991small}
X.~Mao, M.~Saito, H.~Takahashi, {\it Scr Metall Mater} \textbf{1991}, {\it 25}
  (11), 2481--2485.

\bibitem{misawa1989fracture}
T.~Misawa, S.~Nagata, N.~Aoki, J.~Ishizaka, Y.~Hamaguchi, {\it J Nucl Mater}
  \textbf{1989}, {\it 169}, 225--232.

\bibitem{dymavcek2009creep}
P.~Dymáček, K.~Milička, {\it Mater Sci Eng A} \textbf{2009}, {\it 510-511},
  444--449.

\bibitem{gulccimen2013determination}
B.~Gülçimen, P.~Hähner, {\it Mater Sci Eng A} \textbf{2013}, {\it 588},
  125--131.

\bibitem{sainte2002small}
C.~Sainte~Catherine, J.~Messier, C.~Poussard, S.~Rosinski, J.~Foulds, {\it AM
  Soc Test Mater} \textbf{2002}, {\it 4}, 350--370.

\bibitem{fleury1998small}
E.~Fleury, J.S. Ha, {\it Int J Press Vessels Pip} \textbf{1998}, {\it 75} (9),
  699--706.

\bibitem{song2012comparison}
M.~Song, K.~Guan, W.~Qin, J.~A. Szpunar, {\it Nucl Eng Des} \textbf{2012}, {\it
  247}, 58--65.

\bibitem{abendroth2016assessment}
M.~Abendroth, S.~Soltysiak \bblin{} {\it Recent Trends in Fracture and Damage
  Mechanics}, Springer International Publishing, \textbf{2015}, \bblpp{}.
  127--157.

\bibitem{abendroth2003determination}
M.~Abendroth, M.~Kuna, {\it Comp Mater Sci} \textbf{2003}, {\it 28} (3-4),
  633--644.

\bibitem{abbassi2013parameter}
F.~Abbassi, T.~Belhadj, S.~Mistou, A.~Zghal, {\it Mater Design} \textbf{2013},
  {\it 45}, 605--615.

\bibitem{huber2001determination}
N.~Huber, A.~Konstantinidis, Ch. Tsakmakis, {\it J Appl Mech} \textbf{2000},
  {\it 68} (2), 218--223.

\bibitem{sangkharat2019using}
T.~Sangkharat, S.~Dechjarern, {\it Proc Manuf} \textbf{2019}, {\it 29},
  390--397.

\bibitem{bradski2008learning}
Gary Bradski, Adrian Kaehler, {\it Learning {OpenCV:} {Computer} vision with
  the {OpenCV} library}, O'Reilly Media, Inc., \textbf{2008}.

\bibitem{hurst2010european}
R.~Hurst, K.~Matocha, {\it Metall J} \textbf{2010}, {\it 63}, 5--11.

\bibitem{kohlar2017gefuge}
S.~Kohlar, {\it Wissenschaftlich-Technische Berichte / Helmholtz-Zentrum
  Dresden-Rossendor} \textbf{2017}, {\it HZDR-082 2017}.

\bibitem{Houska2017}
M.~Houska, E.~Altstadt, {\it Eur Commiss JRC} \textbf{2017}, {\it SP tests on
  P92 (T70175) at RT Version 1.0}.

\bibitem{Heintze2009ion}
C.~Heintze, C.~Recknagel, F.~Bergner, M.~Hernández-Mayoral, A.~Kolitsch, {\it
  Nucl Instrum Methods Phy Res B} \textbf{2009}, {\it 267} (8-9), 1505--1508,
  Proceedings of the 16th International Conference on Ion Beam Modification of
  Materials.

\bibitem{Tavassoli2004materials}
A.-A.F Tavassoli, A.~Alamo, L.~Bedel, L.~Forest, J.-M. Gentzbittel, J.-W.
  Rensman, E.~Diegele, R.~Lindau, M.~Schirra, R.~Schmitt, H.C. Schneider,
  C.~Petersen, A.-M. Lancha, P.~Fernandez, G.~Filacchioni, M.F. Maday,
  K.~Mergia, N.~Boukos, Baluc, P.~Spätig, E.~Alves, E.~Lucon, {\it J Nucl
  Mater} \textbf{2004}, {\it 329-333}, 257--262, Proceedings of the 11th
  International Conference on Fusion Reactor Materials (ICFRM-11).

\bibitem{Zurbuchen2009influence}
C.~Zurbuchen, {\it ASME Pressure Vessels and Piping Conference} \textbf{2009},
  {\it Volume 5: High Pressure Technology; Nondestructive Evaluation Division;
  Student Paper Competition}, 511--517.

\bibitem{Viehrig2006application}
H.~W. Viehrig, M.~Scibetta, K.~Wallin, {\it Int J Press Vessels Pip}
  \textbf{2006}, {\it 83} (8), 584--592.

\bibitem{dymavcek2013investigation}
P.~Dym{\'a}{\v{c}}ek, F.~Dobe{\v{s}}, Petr Kr{\'a}l, J.~Dvo{\v{r}}{\'a}k,
  \bblin{} {\it Acta Metallurgica Slovaca-Conference}, \textbf{2013}, \bblpp{}.
  57--64.

\bibitem{kumar2015evaluation}
K.~Kumar, A.~Pooleery, K.~Madhusoodanan, R.N. Singh, J.K. Chakravartty, R.S.
  Shriwastaw, B.K. Dutta, R.K. Sinha, {\it J Nucl Mater} \textbf{2015}, {\it
  461}, 100--111.

\bibitem{bruchhausen2016recent}
M.~Bruchhausen, S.~Holmström, I.~Simonovski, T.~Austin, J.-M. Lapetite,
  S.~Ripplinger, F.~de~Haan, {\it Theor Appl Fract Mech} \textbf{2016}, {\it
  86}, 2--10.

\bibitem{james2013introduction}
G.~James, D.~Witten, T.~Hastie, R.~Tibshirani, {\it An Introduction to
  Statistical Learning}, {\it \bblvol{} 112}, Springer New York, \textbf{2013}.

\bibitem{jain2016new}
A.~Jain, G.~Hautier, S.P. Ong, K.~Persson, {\it J Mater Res} \textbf{2016},
  {\it 31} (8), 977--994.

\end{thebibliography}

\end{document}